\documentclass[10pt,journal,compsoc]{IEEEtran}
\newif\ifpeerreview

\peerreviewfalse

\usepackage[nocompress]{cite}
\usepackage{url}
\usepackage{amsmath,amssymb,graphicx}
\usepackage{makecell}

\usepackage{lipsum} 

\usepackage[switch]{lineno}
\usepackage[table,xcdraw]{xcolor}

\usepackage{algorithm}
\usepackage{algpseudocode}
\usepackage{tabularx}
\usepackage{booktabs}
\usepackage{array}

\usepackage[symbol]{footmisc}
\usepackage{xurl}     
\usepackage{hyperref}
\hypersetup{
    colorlinks=false
}

\newcommand*{\phisig}{\ensuremath{\Phi_\text{sig}} }
\newcommand*{\phibkg}{\ensuremath{\Phi_\text{bkg}} }
\newcommand*{\PhiHatsig}{\ensuremath{\widehat{\Phi}_\text{sig}} }
\newcommand*{\PhiHatsiginit}{\ensuremath{\widehat{\Phi}_\text{sig\_init}} }
\newcommand*{\PhiHatbkg}{\ensuremath{\widehat{\Phi}_\text{bkg}} }
\newcommand*{\Tr}{\ensuremath{T_\mathrm{r}} }

\def\argmin{\mathop{\mathrm{arg\,min}} }

\newcommand{\paperID}{29}
\newcommand{\mytitle}{High-Flux Count-Free\\
Single-Photon 3D Cameras}

\title{\mytitle}

\author{Kaustubh~Sadekar,
        Vivek~K~Goyal,
        David~Maier,
        and~Atul~Ingle
\IEEEcompsocitemizethanks{\IEEEcompsocthanksitem K. Sadekar, D. Maier, and A. Ingle are with Portland State University.\protect\\
E-mail: \{ksadekar, maier, ingle2\}@pdx.edu
\IEEEcompsocthanksitem V. K. Goyal is with Boston University.
Email: goyal@bu.edu
\IEEEcompsocthanksitem Project page: \url{https://kaustubh-sadekar.github.io/High-Flux-Count-Free-Single-Photon-3D-Cameras/}.}
}

\begin{document}

\IEEEtitleabstractindextext{%
\begin{abstract}
Single-photon cameras based on single-photon avalanche diode (SPAD) technology are gaining popularity for 3D sensing, thanks to their extreme sensitivity and time resolution.
There are two key challenges with single-photon cameras that limit their widespread use: (i) they suffer from non-linear distortions called ``pile-up'' when operated in high-photon-flux conditions, and (ii) they generate a large volume of raw photon data, creating a severe data bottleneck at each sensor pixel.
In this work, we show that while compressive capture techniques successfully mitigate data transfer challenges, they exacerbate the effects of dead-time distortion because they fail to retain sufficient information about the photon detection history to allow post-processing pile-up correction via existing methods.
We propose a new computational-imaging method that combines free-running capture with an analysis-by-synthesis software pipeline to mitigate pile-up distortions.
Our results with hardware emulations and full-scene and single-pixel simulations show that our method can reliably capture scene distance and reflectance over a wide range of illumination conditions.
Our work will enable high-resolution SPAD cameras that are severely bandwidth-constrained to operate in real-world high-flux scenarios.
\end{abstract}

}

\maketitle

\IEEEraisesectionheading{
  \section{Introduction}\label{sec:intro}
}
Single-photon cameras (SPCs) based on single-photon avalanche diode (SPAD) arrays are an increasingly popular choice for 3D sensing in memory- and bandwidth-constrained settings, such as consumer devices \cite{st1billion}, smartphones \cite{yoshida2021}, and autonomous vehicles \cite{ouster}.
SPCs capture 3D scene information by recording the round-trip time-of-flight of a short light pulse for each scene point, constructing a histogram of photon arrival times for each corresponding pixel, and estimating distance from the histogram peak.
However, increasing spatial resolution creates a severe data bottleneck.
A megapixel SPAD camera capturing 1000-bin histograms at 30 frames per second generates over 30 GB of raw data per second.
To circumvent this bottleneck, recent efforts have explored ways of compressing this histogram data on the fly \cite{gutierrez2022compressive,gutierrez2023learned}, summarizing the data using statistical techniques \cite{sheehan2021sketching,sheehan2024spline,tontini2023histogram}, and using alternative compressed representations called \emph{equi-depth histograms} \cite{countfree2023,pedh2024}.
These methods sacrifice the full photon timestamp history in favor of more parsimonious, lossy representations to fit stringent bandwidth limits.
\begin{figure}[t]
    \centering
    \includegraphics[width=1\linewidth]{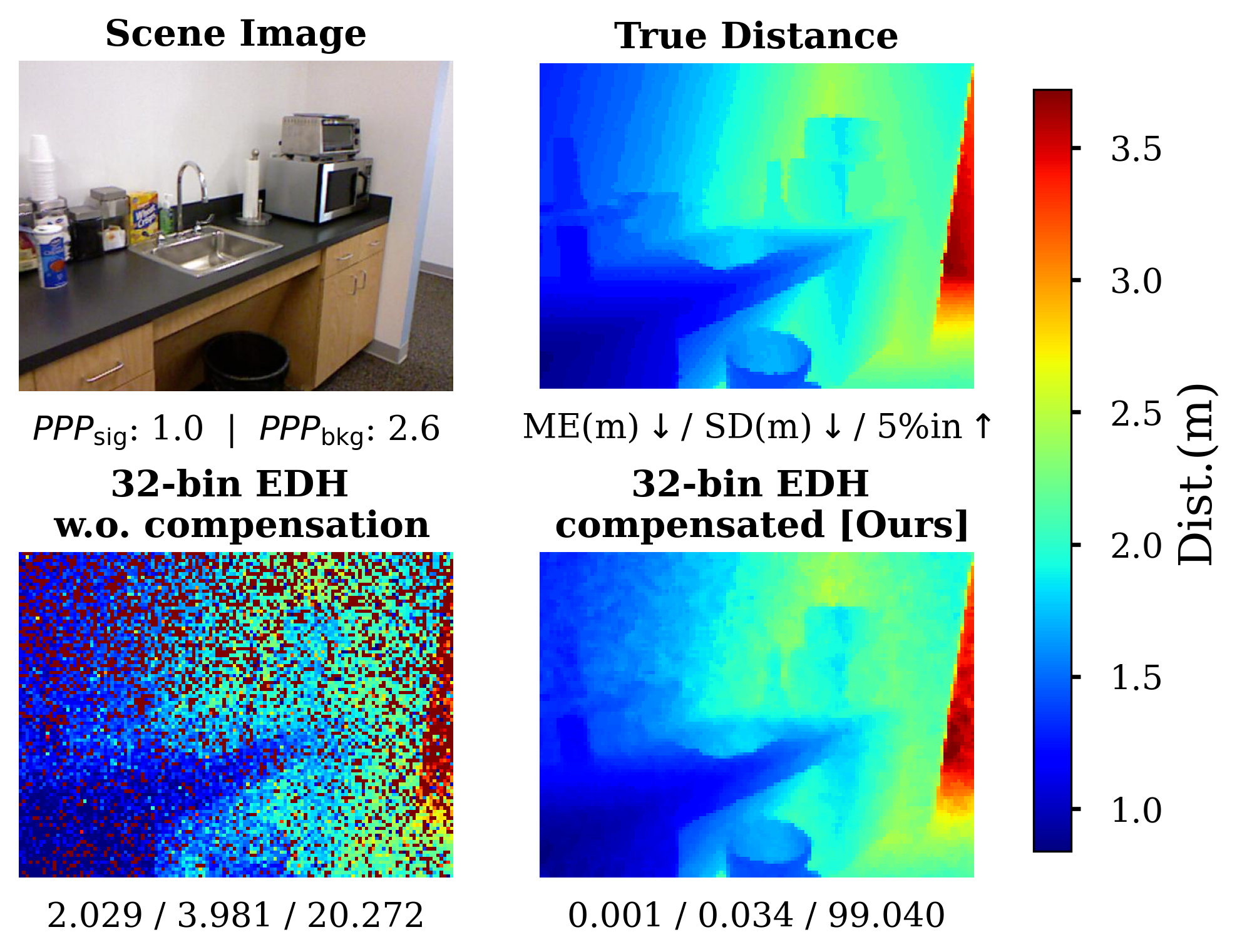}
    \vspace{-20px}
    \caption{\textbf{Our method enables dense 3D imaging with dead-time-affected SPCs in memory-constrained settings and high-photon-flux conditions.} Without compensation, uncorrected distance estimates (bottom left) suffer from dead-time distortion and lossy compression via a count-free histogrammer. In contrast, our approach extracts depth directly from these compressed, pile-up-distorted measurements, yielding dense 3D maps with orders-of-magnitude higher accuracy. (Bottom numbers report mean error [ME], standard deviation [SD], and 5\% inliers).
    }
    \vspace{-20px}
    \label{fig:teaser}
\end{figure}

\begin{table}
\centering
\scalebox{0.94}{
\begin{tabular}{@{}rccc}
\toprule
\textbf{Method} & \textbf{\begin{tabular}[c]{@{}c@{}}Pixel\\ Resources\end{tabular}} & \textbf{\begin{tabular}[c]{@{}c@{}}Robust\\ to Pile-up\end{tabular}} & \textbf{\begin{tabular}[c]{@{}c@{}}High-res. \\ Hardware\\ Compatible\end{tabular}} \\ \midrule
High-flux SPL \cite{rapp2021high} & \cellcolor[HTML]{FFCCC9}High & \cellcolor[HTML]{32CB00}Yes & \cellcolor[HTML]{FFCCC9}No \\
CSPH \cite{gutierrez2022compressive,gutierrez2023learned} & \cellcolor[HTML]{32CB00}Low & \cellcolor[HTML]{FFCCC9}No & \cellcolor[HTML]{32CB00}Yes \\
Hist.-less \cite{tontini2023histogram} & \cellcolor[HTML]{32CB00}Low & \cellcolor[HTML]{FFCCC9}No & \cellcolor[HTML]{32CB00}Yes \\
SplineSketch \cite{sheehan2024spline} & \cellcolor[HTML]{32CB00}Low & \cellcolor[HTML]{FFCCC9} No$^a$ & \cellcolor[HTML]{FFCCC9}No \\
EDH \cite{countfree2023,pedh2024} & \cellcolor[HTML]{32CB00}Low & \cellcolor[HTML]{FFCCC9}No & \cellcolor[HTML]{FFCCC9}No \\ \midrule
This Work & \cellcolor[HTML]{32CB00}Low & \cellcolor[HTML]{32CB00}Yes & \cellcolor[HTML]{32CB00}Yes \\ \bottomrule
\end{tabular}
}
\vspace{0.05in}
\caption{\protect\parbox[t]{\linewidth}{\textbf{Comparison of state-of-the-art.} Although many existing techniques deal with the problems of SPC data compression and pile-up separately, there is no single technique that solves these problems jointly and is amenable to in-sensor implementation for dense pixel arrays. ($^a$The SplineSketch method handles signal peak pile-up, but only over a narrow range of signal flux levels using a look-up table).}}\label{tab:comparison}
\vspace{-20px}
\end{table}

In high-flux conditions, SPC pixels suffer from "dead-time," a reset period of tens to hundreds of nanoseconds after each detection that causes earlier-arriving photons to be preferentially detected, resulting in significant distortions in the detected photon stream.
This non-linear distortion, known as ``pile-up,'' dwarfs and shifts the true signal peak, leading to biased distance estimates.
Fig.~\ref{fig:teaser} shows a simulated indoor scene captured by a compressive technique \cite{countfree2023} at an average photons-per-pixel (PPP) flux 300 times higher than the 5\% rule of thumb, which is frequently used to avoid SPAD pile-up \cite{wahl2014tcspc}.

Existing pile-up mitigation techniques rely on a combination of hardware and computational strategies (summarized in Table~\ref{tab:comparison}). Hardware approaches use optical \cite{gupta2019photon} or electrical filtering \cite{tontini2022comparison}, or introduce temporal shifts between the laser and detector \cite{gupta2019asynchronous,rapp2021high}.
However, it is unclear whether these hardware approaches sufficiently reduce pile-up distortions to allow compressive approaches to successfully recover scene distance.
Computational approaches use probabilistic modeling and statistical estimation to invert pile-up. However, these methods require high-resolution histograms to preserve the full temporal history of photon arrivals. Consequently, they cannot be applied to compressive techniques that discard this granular data in favor of a parsimonious, lossy representation.
Here we ask: Is it possible to obtain reliable distance estimates from compressive SPCs even in the presence of strong pile-up distortions?

\smallskip
\noindent
\textbf{Scope and Limitations:}
Although there are several techniques to compressively capture SPAD data for 3D sensing, here we limit ourselves to a specific kind of compressive capture technique that relies on ``equi-depth'' photon histograms\footnote[2]{The word ``depth'' in equi-depth histograms refers to the counts in each histogram bin and should not be confused with scene distance. To avoid confusion, we use the term ``distance map'' throughout this paper. While ``equal-height'' is more intuitive, we choose ``equi-depth'' to maintain consistency with terminology used in prior work \cite{countfree2023}.}
We propose a model-based analysis-by-synthesis algorithm to estimate scene distance and reflectance while compensating for dead-time distortions. We evaluate our approach through extensive single-pixel and full-scene simulations, alongside hardware emulation studies, over a wide range of illumination conditions and dead-time values.
\section{Related Work}\label{sec:relatedwork}

\smallskip
\noindent
\textbf{SPCs for 3D Sensing:}
Early methods focused on 3D SPCs that used a single-pixel SPAD and raster-scanned the scene \cite{gupta2019photon, kirmani2014first, buller2007ranging, pawlikowska2017single}.
Most existing approaches rely on single-pixel detectors, or linear arrays \cite{lindell2018sensor, spadnet}, while some other studies have used low-resolution arrays of 32\,$\times$\,32 SPAD pixels \cite{shin2016photoneff, 5587099}.
Due to the large memory requirement of existing pile-up mitigation techniques that employ high-resolution photon histograms or timestamp streams, existing SPCs trade off spatial pixel resolution for distance resolution to maintain manageable data rates.
Our work can potentially avoid this tradeoff by capturing a more succinct photon stream representation in-pixel while also maintaining robustness to pile-up.

\smallskip
\noindent
\textbf{Resource-constrained SPCs: } Improvements in SPC hardware designs have been proposed to lower data rates and power consumption. 
Most methods rely on constructing an \emph{equi-width (EW) histogram}. 
However, unlike conventional SPCs, these designs can process in a coarse-to-fine fashion by iteratively zooming into the region-of-interest \cite{tcspc2, zooming2, zooming3, adaptivegating}, or can share the histogramming circuit resources on the sensor \cite{resourceshare}. 
Compressing photon timestamps on the fly using different coding techniques has been shown in simulations to provide more than a 10-fold reduction in bandwidth  \cite{gutierrez2022compressive, Tachella2022SketchedRH}.
However, additional information, such as a coding matrix \cite{gutierrez2022compressive}, must be stored and simultaneously accessed by all the pixels, which is challenging to implement using existing SPC designs. In contrast, researchers have proposed resource-efficient SPC designs that are more amenable to hardware implementation \cite{pedh2024, countfree2023}. 
Unfortunately, none of the existing work considers the problems of pile-up mitigation and compression jointly.
Here we show that a free-running SPC pixel \cite{gupta2019asynchronous,rapp2021high} coupled with an \emph{equi-depth histogrammer} \cite{countfree2023,pedh2024} can simultaneously achieve memory-efficiency and robustness to pile-up.

\smallskip
\noindent
\textbf{Dead-time Compensation:} 
Existing methods for dealing with dead-time-induced pile-up rely on a combination of hardware approaches and computational postprocessing.
Computational methods use probabilistic models to invert the effect of pile-up on the photon-detection histograms \cite{coates1968correction}.
Some of the initial methods were designed assuming SPCs that can only detect a single photon per laser cycle \cite{synchronous-model1, sync-dtc1, sync-dtc2}.
More recent methods considered advanced SPCs, which can detect multiple photons per laser cycle \cite{rapp2021high}. 
These methods require either a complete history of photon detection timestamps or high-resolution equi-width histograms.
Recent sketching-based approaches can reduce signal peak pile-up while maintaining low memory footprint, but these look-up-table-based methods only work over a narrow range of signal flux levels \cite{sheehan2024spline}.
In contrast, our method handles a wide range of both background and signal flux conditions.
In this work, we take inspiration from previous methods \cite{rapp2019dead,rapp2021high,gupta2019asynchronous} and derive a probabilistic model for the effect of dead-time on equi-depth histogrammers, and propose an analysis-by-synthesis algorithm to remove pile-up distortions.
By avoiding the need to store high-resolution photon timestamp streams, our method can operate in severely memory-constrained or bandwidth-constrained settings.
\section{SPC Preliminaries}\label{sec:prelims}

\begin{figure}[t]
    \centering
    \includegraphics[width=1\linewidth]{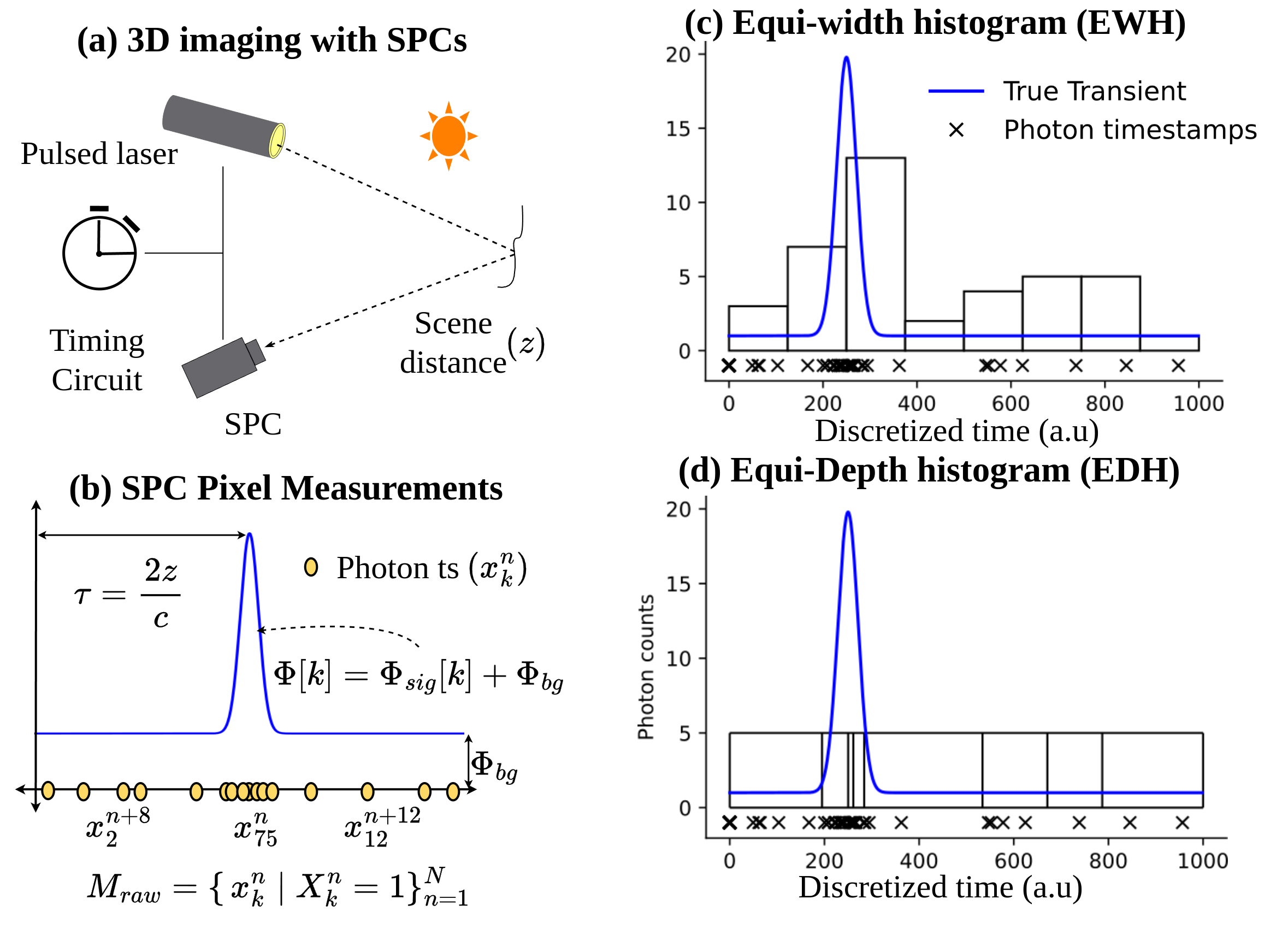}
    \vspace{-15px}
    \caption{\textbf{SPC image formation and memory-efficient histogramming techniques.} (a) Each SPC pixel operates in tandem with a pulsed laser whose round-trip travel time gives an estimate of the scene distance. 
    (b) The pixel captures the returning photon events as a stream of photon timestamps and attempts to locate the true peak location in the true underlying photon distribution $\Phi[t]$. 
    (c) High-resolution timestamp data consumes too much in-pixel memory. One way to compress it is to use an equi-width histogram with very few bins (8 bins in this case).
    These wide bins have limited distance resolution and introduce quantization artifacts.
    (d) An equi-depth histogram with 8 bins is a more efficient representation to track the laser-peak location as seen from a small cluster of bin boundaries around the true peak location.
    (Figure adapted from \cite{pedh2024}.)}
    \label{fig:img_model}
\end{figure}

\subsection{Light Transport and Detection Model}
An SPC pixel (see Fig.~\ref{fig:img_model}(a)) operates in concert with a pulsed laser that illuminates each scene point with a periodic pulse train with a repetition period $T_r$.
Photon return events are time-tagged by each pixel, where the time is measured ``modulo-$T_r$'' with respect to the most recently emitted laser pulse.
The detected photon stream contains both the laser photons reflected from the scene point of interest and background illumination due to other ambient sources of light in the scene.
For $t \in [0,T_r)$, the time-varying light intensity $\Phi(t)$ incident on the SPC pixel can be written as $\Phi(t) = \phisig \, g(t-t_0) + \phibkg$ where $\phisig$ denotes the laser photon flux, $g(\cdot)$ encodes the laser pulse shape, and $\phibkg$ denotes the background photon flux.
Scene effects such as reflectance, albedo, and distance-squared, cosine-fourth falloffs are all absorbed in the $\phisig$ term.
The $\phibkg$ term also includes other sources of SPC pixel noise (e.g., dark noise) that are uncorrelated with the laser photons.
We assume a Gaussian pulse shape $g(\cdot)$ where the pulse standard deviation is known (from a datasheet, or perhaps through prior calibration captures).
In an ideal case of no background illumination and an infinitesimally narrow laser pulse width, a single laser photon detection would be enough to recover the distance of a scene point.
The photon's detection time $t_0$ is related to the scene point's distance through $z_0 = c t_0 / 2$, where $c$ is the speed of light.

In practice, it is necessary to capture many photons over multiple laser cycles to recover an accurate distance estimate.
Assuming a timestamp resolution of $\Delta t$, each SPC pixel captures photon arrival timestamps that lie in one of $B = {\Tr}/{\Delta t}$ discrete window locations.
In a low-photon-flux scenario and absence of multiple reflections, the mean number of photons received by the SPC pixel in the $k^\text{th}$ window location is given by:
\begin{equation} \label{eq:rn_full} 
\Phi[k] =  \phisig \int_{k \Delta t}^{(k+1) \Delta t} g\left(t - t_0\right) \,dt \ +  \phibkg \Delta t.
\end{equation}
We call $\Phi[k]$ the \emph{true transient distribution}.
We define the \textit{signal-to-background ratio} (SBR) as $\mathsf{SBR} \overset{\operatorname{def}}{=} \phisig / \phibkg$.
The number of photons received in any window location $k$ follows a Poisson distribution.
Typically, an SPC pixel can detect at most one photon per window location in any given laser cycle.
Moreover, after each photon-detection event, the pixel needs a finite amount of time, called the \emph{dead-time}, to reset before it can detect the next photon.
The stream of photon detection events $\left(X^n_k\right)_{k=0}^{B-1}$ in the $n^\text{th}$ laser cycle is defined recursively in terms of the previous detection events: $X^n_k = 1$ if at least one photon was incident in window location $k$ in that cycle, and $X^n_i = 0$ for $k-t_d \leq i \leq k-1$ where $t_d$ is the dead-time (in discrete time units).
This photon loss due to dead-time causes a non-linear pile-up distortion.
The statistical average of these photon detection events---the \emph{perceived transient distribution}---does not equal the true transient distribution described by Eq.~(\ref{eq:rn_full}).
Photon streams are usually captured over $N_r$ laser cycles, usually in the hundreds to thousands.
The collection of all photon detections $(X^n_k)_{0\leq k \leq B-1, 0 \leq n \leq N_r-1}$ is called the \emph{photon data cube}  and requires a large volume of on-sensor memory for a focal plane array of kilo-to-megapixel resolution.

\subsection{Conventional Histogram Formation}

\begin{figure}
    \centering
    \includegraphics[width=1\linewidth]{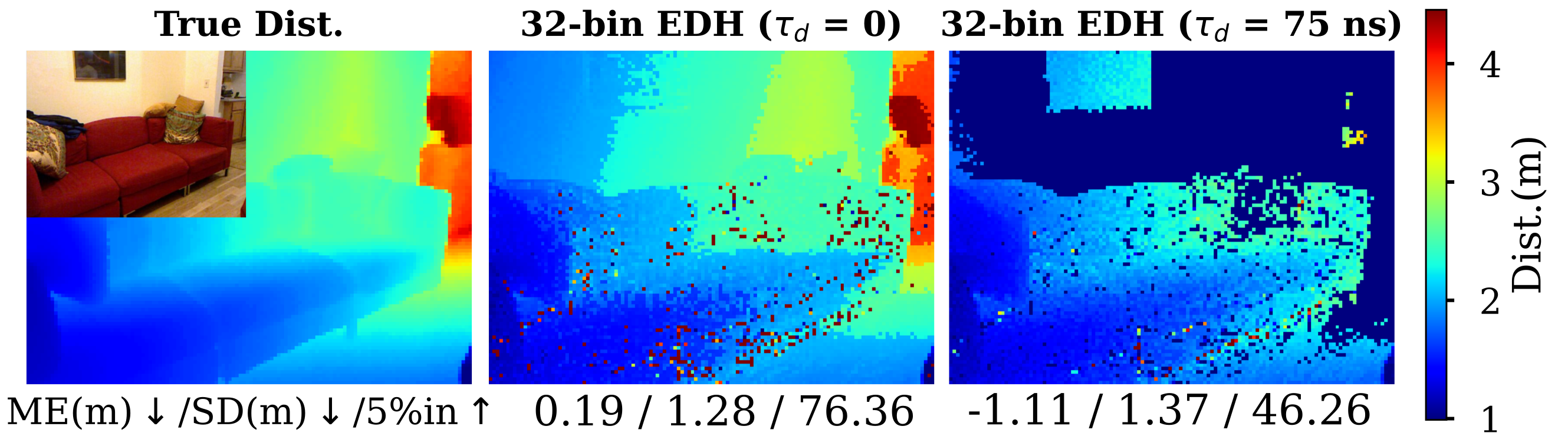}
    \vspace{-15px}
    \caption{\textbf{Effect of dead-time distortion on EDH SPCs.} In the absence of dead-time distortions, a compressive capture from a 32-bin EDH provides reliable estimates of the true scene distances.
    For a realistic dead-time of $\tau_d=75$\,ns, the scene distance estimates are severely corrupted by pile-up.
    Due to the non-linearity of pile-up distortions, statistical averaging by increasing the capture time does not help.}
    \label{fig:dead_time_edhspc}
\end{figure}

A conventional SPC pixel must repeatedly sample hundreds to thousands of photon timestamps to reconstruct the true peak of the distribution, corresponding to the laser pulse, to estimate the round-trip time-of-flight.
Due to hardware constraints, it is impossible to practically process this raw photon data in the sensor or transfer it to any off-sensor compute module, which motivates the need for techniques to compress the raw measurements.

\smallskip
\noindent \textbf{Conventional Equi-width (EW) Histograms:}
The most widely used processing technique for SPCs is to summarize the photon data cube into an equi-width (EW) photon-count histogram for each pixel.
The histogram maintains a count of photon detections over equally spaced time bins, spanning the laser period $T_r$.
The peak of this EW histogram serves as an estimate of the time-of-flight and hence the scene distance.
Although histogramming does reduce the size of the raw photon data cube, it still requires more memory than can be stored in-sensor.
In practice, these EW histograms require $\approx 1000$ bins per pixel.
For recent SPAD-based SPCs that have approached megapixel resolutions, the amount of EW histogram data adds up to several gigabytes per second when operating at video rates.
For SPCs to become mainstream, especially for resource-constrained applications, such as low-power mobile robots and smartphone cameras, it is desirable to reduce the volume of the photon data cube without compromising the spatial or temporal resolution of the final distance maps.

\smallskip
\noindent
\textbf{Compressive Techniques:} 
A straightforward method to reduce the memory requirement is to reduce the number of EW histogram bins.
The low number of equi-depth bins causes severe quantization artifacts in the estimated scene distance.
An example is shown in Fig.~\ref{fig:img_model}(c) with eight EW bins.
Other methods for compressive capture rely on capturing a lower-dimensional representation of the high-resolution EW histogram using linear projections \cite{gutierrez2022compressive} or sketching \cite{sheehan2021sketching}.
Recently, ``equi-depth'' (ED) histograms were proposed as a more resource-efficient alternative to EW histograms for capturing ``peaky'' transient distributions such as that of a laser pulse \cite{countfree2023}.
As shown in an example in Fig.~\ref{fig:img_model}(d), an ED histogram uses variable-width bins such that each bin contains an equal number of photon counts.
In this work, we focus on ED histograms and propose a method for compensating for dead-time distortions on ED histogram measurements.

\subsection{Equi-depth Histogrammers for SPCs}\label{subsec:edhspcs}
ED-histogram bin boundaries can be estimated using an \textit{equi-depth histogrammer (EDH)} in an online fashion, without the need to store the entire photon data cube.
An EDH operates on a per-pixel basis, capturing ED bin boundaries, usually consisting of equally spaced quantiles.
The basic building block of an EDH is a \emph{binner circuit} \cite{countfree2023, pedh2024} that tracks any arbitrary quantile of the incident transient distribution.
The \emph{control value (CV)} of the binner is updated iteratively, at each photon detection event.
In general, the $\text{CV}$ is decremented (incremented) by a step if the most recent photon detection is earlier (later) than the current $\text{CV}$.
The relative sizes of the increment and decrement steps enable tracking arbitrary quantiles (e.g., a 1:1 ratio tracks the median, while a 3:1 ratio tracks the $75^\text{th}$ percentile).
In its steady state, the binner's CV converges (probabilistically) to the target quantile.
Thus, an EDH can be thought of as a \emph{quantile sensor} that directly captures the quantile positions of a transient distribution.

\smallskip
\noindent
\textbf{Limitation of Previous EDH Designs:} Researchers have proposed various stepping-strategies to speed up binner convergence and reduce the variance in the CV \cite{countfree2023, pedh2024}. 
However, they assume a linear forward model, which implies that the SPC pixel detects every photon incident on the pixel.
In practice, when operating in high-flux scenarios, the detected photon stream will suffer from severe pile-up distortions.
The probability of detecting later-arriving photons is reduced due to dead-time. 
The seriousness of pile-up distortion is shown in an example in Fig.~\ref{fig:dead_time_edhspc}.
Observe how the mean distance estimate suffers from a strong negative bias in the high-flux scenario.

In the next section, we derive a forward imaging model for EDH SPCs operating in high-flux conditions and design an improved distance estimator that mitigates pile-up distortions.
A key advantage of our approach is that it does not require any modifications to existing EDHs; we can use them as designed and compensate for dead-time distortions after capture of the EDH boundaries.
\begin{figure}
    \centering
    \includegraphics[width=1\linewidth]{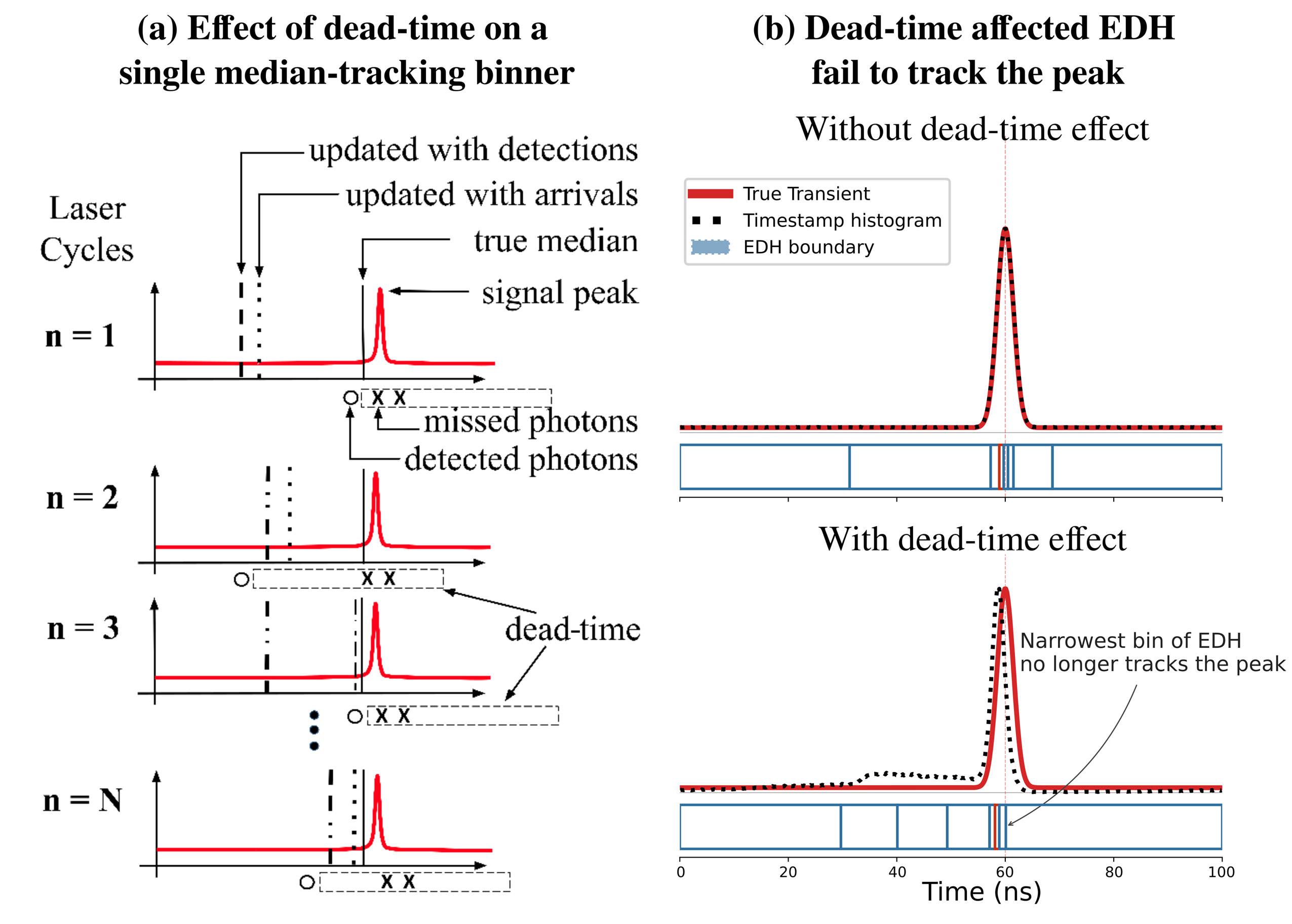}
    \vspace{-15px}
    \caption{\textbf{Effect of dead-time on SPC measurements.} (a) In case of no dead-time, the binner tracks the median of the photon arrivals, but in case of dead-time, the binner tracks just the detections and suffers from a negative bias due to dead-time distortion. (b) When dead-time = 0, the narrowest bin tracks the peak, but in the presence of dead-time, the narrowest bin no longer tracks the peak of the photon arrivals.}
    \label{fig:binner-update-vs-deadtime}
\end{figure}

\begin{figure*}[!th]
    \centering
    \includegraphics[width=\linewidth]{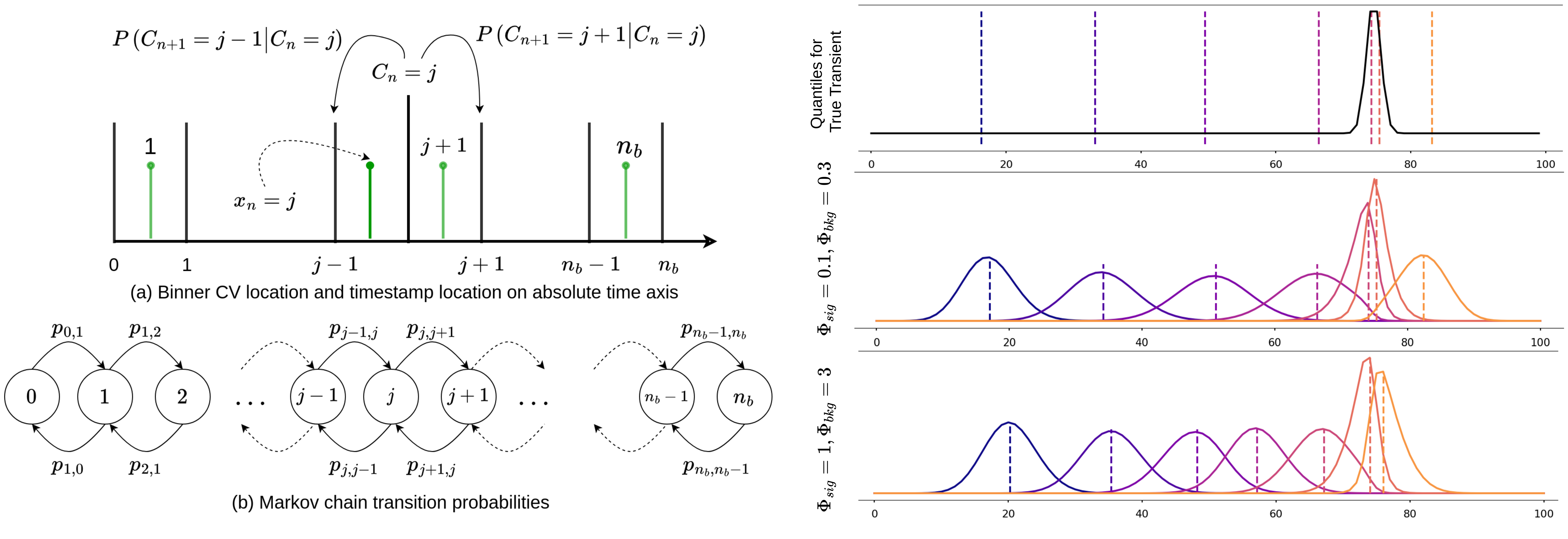}
    \vspace{-5px}
    \caption{
    \textbf{Markov chain model for an equi-depth histogrammer.}
    (a) The photon detection stream lives on a discrete state space with $B$ possible locations; the binner control value lives on a state space of $B+1$ possible locations.
    The discrete photon detection times can be described using a Markov chain \cite{rapp2021high}.
    (b) A binner's control value is incremented or decremented with some probability that is tied to the photon detection probabilities relative to the current control-value position.
    The binner's control value, therefore, follows another Markov chain, which is driven by the Markov chain of photon detections.
    (c) An EDH consists of a bank of binners, each tracking a different quantile (shown by vertical dashed lines) of the perceived transient distribution, which may be different from the true transient distribution due to dead-time distortion.
    The binners converge to their respective stationary distributions that are centered around the actual quantiles.}
    \vspace{-0.1in}
    \label{fig:binner-markovchain}
\end{figure*}

\section{Scene Recovery from Dead-time-distorted Equi-depth Histograms}\label{sec:deadtime_edh}

Our proposed compressive SPC technique follows a hardware-software co-design approach.
\textit{Hardware:} capturing count-free ED histograms using a ``free-running'' SPC pixel; \textit{Software:} an analysis-by-synthesis computational framework to estimate scene distances from the ED histogram boundary measurements.
Most commercially available SPCs today operate in ``synchronous'' mode, where the start of the photon capture is synchronized with the transmission time of each laser pulse.
When operating in high-flux conditions, this synchronous capture approach exacerbates pile-up, especially due to background photons that form the overwhelming majority of the early detections.
Recent work has shown the benefit of operating SPCs in a free-running mode \cite{gupta2019asynchronous,rapp2019dead,rapp2021high,kitichotkul2025freerunning} to reduce the pile-up distortions due to background photons.

\smallskip
\noindent
\textbf{Motivation to Model Dead-time-distorted Histogrammers:}
Although operating SPCs in a free-running mode mitigates the pile-up distortion due to background photons and improves the chances of detecting a true signal photon, the probabilistic distribution of photon detection times is not identical to that given by the true transient distribution in Eq.~(\ref{eq:rn_full}).
As shown in Fig.~\ref{fig:binner-update-vs-deadtime}, there is still some residual pile-up distortion from the laser signal photons that appears as a ``peak shadow.'' Hence, even if the photon detections were to be accumulated over a long exposure time, due to dead-time, the perceived transient distribution would be different from the true transient distribution.
Hence, simply operating an EDH with photon detection streams of a free-running SPC pixel does not guarantee reliable scene-distance estimates because it tracks the quantiles of the perceived transient, not the true transient.

In this section, we propose a theoretical model for the ED bin boundaries measured by an EDH.
Our work builds on the Markov chain model proposed by Rapp \emph{et al.} \cite{rapp2019dead} that provides a probabilistic model of the photon detection stream from the perceived transient.
Such a model allows us to, in turn, derive a probabilistic model for the quantile estimates generated by an arbitrary quantile-tracking binner.
These probabilistic models ultimately allow us to estimate scene properties through maximum-likelihood estimation in an analysis-by-synthesis approach.

\subsection{EDH Imaging Model with Dead-time \label{sec:trans_mat}}
The photon detection times $\{X_\ell\}_{\ell \in N}$  in a free-running SPC pixel with dead-time can be modeled as a Markov chain with continuous-valued state space $[0, T_r)$ \cite{rapp2019dead}.
We discretize the state space into $B$ uniformly-spaced locations $\{x_k\}^{B}_{k=1}$.
The transition probabilities are given by \cite{rapp2019dead}:
\begin{equation}
f_{X_{\ell+1}\mid X_{\ell}}(x_{\ell+1}\mid x_{\ell})
=
\frac{\Phi\bigl(x_{\ell+1}\bigr)}{1 - e^{-\Phi}}
e^{- \int_{A}^{B}
  \Phi(\tau)\,\mathrm{d}\tau},
\end{equation}
where
\begin{align*}
A &= {x_{\ell} + x_{d}},& B = { 
      \Bigl\lceil\frac{x_{\ell}+x_{d}-x_{\ell+1}}{T_{r}}\Bigr\rceil\,T_{r} \;+\; x_{\ell+1}
    }.
\end{align*}
Here $\Phi(x_k)$ is the Poisson process intensity at detection time $x_k$,
$\Phi$ is the rate of total photons detected per laser cycle, and
$x_d = \tau_d \mod T_r$.
The $B \times B$ Markov transition matrix $P$ over the discrete state-space with has entries: $P_{m,n} := f_{X_{\ell+1}\mid X_{\ell}}\bigl(x_n \mid x_m\bigr)$.
Here the perceived transient is the stationary distribution of $P$ and is denoted by $\widetilde{\Phi}.$
We compute it numerically by calculating the left eigenvector of $P$ corresponding to an eigenvalue of $1$.
Note that in the absence of any dead-time distortions, we expect $\widetilde{\Phi}$ to be identical to the true transient distribution $\Phi$.

The binners that constitute the EDH are driven by a stream of photons that follow a Markov process with a steady-state distribution given by its perceived transient.
We will assume that the Markov chain describing the detected photon stream has reached its steady state, given by this perceived transient.
Next, we derive a mathematical model for a binner circuit that tracks an arbitrary quantile of this detected photon stream.
We show that the binner's control value can be modeled as another Markov chain, which is driven by the Markov chain of photon detection events.

Let $C_{\ell}$ denote an arbitrary quantile-tracking binner's control value (CV) after the $\ell^\text{th}$ photon detection event.
This CV lives on a discrete grid with $B+1$ locations, which is one longer than the discrete grid of photon-detection-event locations, as shown in Fig.~\ref{fig:binner-markovchain}.
The CV is incremented if the next photon detection event occurs at a position that appears later than the CV position and decremented if it appears earlier than the CV position.
There is also a nonzero probability that the photon detection coincides with the current CV, in which case the CV remains unchanged.
Let $p_{i,j}$ denote the CV's transition probability from state $i$ to $j$, where the states are the photon-detection-event locations.
The Markov chain transition matrix for a binner tracking the $q^\text{th}$ quantile is a $(B+1)\times(B+1)$ tridiagonal matrix.
The off-diagonal entries are given by
$p^{(q)}_{j, j-1} = (1-q)\sum^{j}_{i=1} \widetilde{\Phi}[i]$ and
$p^{(q)}_{j, j+1} = q\sum^{B}_{i=j+1} \widetilde{\Phi}[i]$,
while the diagonal elements are defined as
$p^{(q)}_{j, j} = 1 - \left(p^{(q)}_{j,j-1} + p^{(q)}_{j,j+1}\right)$.

We can numerically compute the stationary distribution $\mathcal{D}_{q}$ for this Markov chain.
An example of the stationary distributions of the seven binners constituting an 8-bin EDH is shown in Fig.~\ref{fig:binner-markovchain}.
These binners track the seven equally-spaced quantiles corresponding to $q = i/8$ for $i\in\{1,\ldots, 7\}$.
Observe that the convergence of binners is probabilistic in the sense that they do not converge to the exact quantile positions of the perceived transient, but to these stationary distributions $\mathcal{D}_q.$
The mode of the stationary distribution always aligns with the true quantile position. (For a theoretical proof, please see \ref{supp:theory}).
We observe that the stationary distributions $\mathcal{D}_q$ are always unimodal, but not always symmetric.
Their spread is more narrow and asymmetric when a quantile is closer to the signal peak, but more diffuse and symmetric when farther away from the peak.
Since there is no simple closed-form expression for these stationary distributions, we must resort to numerical simulations to understand the effects of scene parameters and dead-time.
Through extensive empirical simulations over various signal, background, and distance values, we observe that the mode of the stationary distribution $\mathcal{D}_q$ always aligns with the true position of the $q^\text{th}$ quantile of the perceived transient.
(See \ref{supp:theory} for details.)
This property means that no matter how narrow or diffuse the stationary distributions are, the highest probability value is at the perceived transient's exact quantile location.
Moreover, due to the law of large numbers, a repeated-sample-and-average measurement procedure of the binner's control value can provide improved estimates of the exact quantile.
\begin{figure}[!th]
    \centering
    \includegraphics[width=0.95\linewidth]{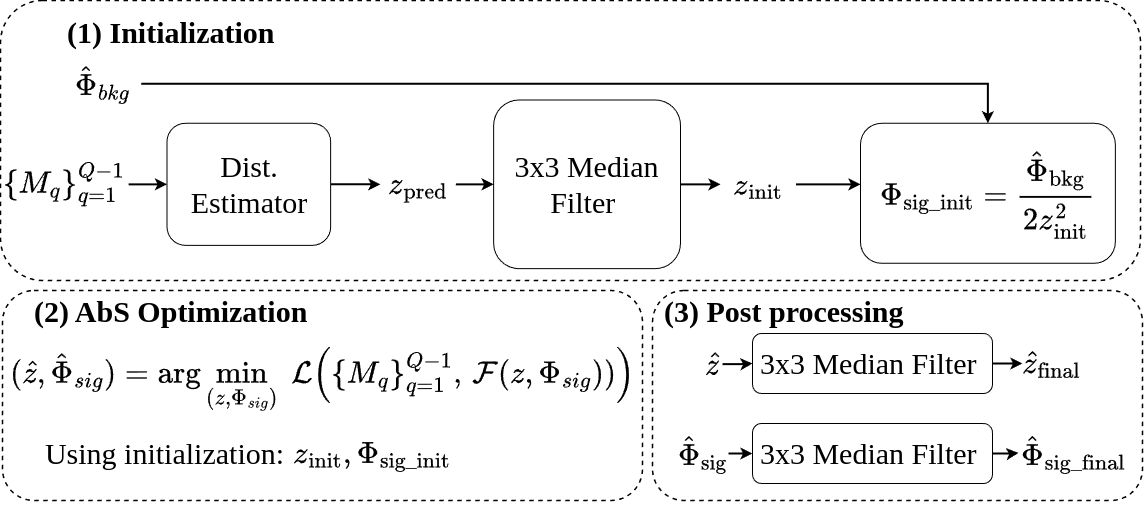}
    \caption{\textbf{Three stages of the analysis-by-synthesis (AbS) pipeline.} (1) Initialization stage uses SPC measurement $M_q$ and background flux  $\PhiHatbkg$ estimated from passive capture. We also apply a spatial $3 \times 3$ median filter to reduce noise in the initial distance estimate. (2) The optimization stage minimizes a loss function to obtain model parameters that best explain the observed EDH data. (3) A postprocessing stage applies a 3x3 median filter on the distance estimate and effective signal estimate to get the final outputs.}
    \vspace{-0.15in}
    \label{fig:abs-pipeline}
\end{figure}

\subsection{Pixel-wise Analysis-by-synthesis Pipeline}\label{subsec:pixel_abs}
The previous section presented a mathematical forward model that describes the output of an EDH driven by a photon stream with dead-time distortions.
We now tackle our main goal---the inverse problem of estimating scene properties (distance and reflectivity) from the EDH measurements.
We propose an analysis-by-synthesis pipeline (Fig.~\ref{fig:abs-pipeline}) which consists of three stages: initialization, optimization, and post-processing.

During the initialization stage, we assume that the background strength for a given pixel is known or can be estimated reliably through a passive capture.
The optimization stage then estimates the scene distance, $\hat{z}$, and signal strength, $\widehat{\Phi}_\text{sig}$, through an iterative process.
During synthesis, we generate a perceived transient by plugging the current estimate $(z, \phisig)$ for the distance and signal strength into the probabilistic forward model from the previous section.
In the subsequent analysis, we compare how well the measured ED histogram boundary values match those synthesized by the forward model.
An optimization routine then updates the current estimate of $(z, \phisig)$.
Denoting the forward model by $\mathcal{F}$, the optimization routine must minimize the difference as measured by a suitable loss function $\mathcal{L}$ between the measured quantile values $\{M_q\}_{q=1}^{Q-1}$ and the quantile values predicted by the forward model $\mathcal{F}$:
\begin{equation}
    (\widehat{z}, \PhiHatsig)
= \argmin_{(z, \phisig)}\;\mathcal{L}\!\left(\{M_q\}_{q=1}^{Q-1},\,\mathcal{F}(z, \phisig)\right). \label{eq:abs_opt}
\end{equation}

\smallskip
\noindent
\textbf{Choice of the Loss Function:}
We use the negative log-likelihood loss function $\mathcal{L}_\text{NLL},$, which considers the inherent stochasticity in the measured CV positions returned by a quantile-tracking binner as modeled by the CV's Markov chain stationary distributions.
Assuming that the binners for different quantiles are run sequentially
(hence, the individual binner CVs can be considered as independent random variables),
the log-likelihood can be written as the product $\prod_{i=1}^{Q-1} \mathcal{D}_q[M_{q}]$, where $\mathcal{D}_q$ is the stationary distribution of the CV tracking the $q^\text{th}$ quantile.
The loss function is simply the negative sum of the log-likelihoods:
$    \ell_\text{NLL} = -\sum_{q=1}^{Q-1} \log \mathcal{D}_q[M_{q}],
$
where $\mathcal{D}_q$ depends on the current $(z, \phisig)$ estimate.
The NLL loss function matches the theoretical model most closely, but suffers from an $O(QB^3)$ computational cost of numerically computing the stationary distributions of the binner CVs at each iterative update of the analysis-by-synthesis optimization loop (for loop in Line 18 of Alg.~\ref{alg:abs_pseudocode}).
In \ref{supp:loss}, we demonstrate results for two other loss functions (boundary-MSE and quantile-MSE) that are approximate proxies for the NLL loss with a much lower $O(Q)$ computational cost (see Line 5 in Alg.~\ref{alg:abs_pseudocode}).
In practice, we found that a ``dual loss'' strategy is beneficial. 
We first minimize the faster-to-compute, but less accurate, boundary-MSE loss and then switch to the slower but more accurate NLL loss.

\algrenewcommand\algorithmicrequire{\textbf{Input:}}
\algrenewcommand\algorithmicensure{\textbf{Output:}}
\begin{algorithm}[t]
\footnotesize
\caption{Analysis-by-Synthesis (AbS) Optimization}
\label{alg:abs_pseudocode}
\begin{algorithmic}[1]
\Require Initial guess $(z_0, \Phi_{\text{sig},0})$, measured EDH boundaries $\{M_q\}_{q=1}^{Q-1}$
\Ensure Refined estimates $(\widehat{z}, \widehat{\Phi}_{\text{sig}})$
\Function{AbS\_Optimize}{$z_0, \Phi_{\text{sig},0}, \{M_q\}_{q=1}^{Q-1}$}
    \Statex \vspace{1mm}\hrule\vspace{1mm}
    \State //\textbf{Coarse Optimization (Proxy Loss)}
    \Function{BMSE\_Loss}{$z, \Phi_{\text{sig}}, \{M_q\}_{q=1}^{Q-1}$}
        \State $\{C_q\}_{q=1}^{Q-1} \leftarrow \mathcal{F}(z, \Phi_{\text{sig}})$ 
        \State $\ell_{\text{BMSE}} \leftarrow \frac{1}{Q-1} \sum_{q=1}^{Q-1} \| M_q - C_q \|^2$
        \State \Return $\ell_{\text{BMSE}}$
    \EndFunction
    \State $(z_1, \Phi_{\text{sig},1}) \leftarrow \text{NelderMead}(\text{BMSE\_Loss}, z_0, \Phi_{\text{sig},0}, \text{maxiter}\!\!=\!\!50)$
    
    \Statex \vspace{1mm}\hrule\vspace{1mm}
    \State //\textbf{Compute $q^\text{th}$ binner's stationary distribution}
    \Function{SolveStatDist}{$z, \Phi_{\text{sig}}, q$}
        \State Construct transition matrix $P_q$ (see Sec.~\ref{sec:trans_mat})
        \State Compute stationary distribution $\mathcal{D}_q$ by solving $\mathcal{D}_q P_q = \mathcal{D}_q$
        \State \Return $\mathcal{D}_q$
    \EndFunction

    \Statex \vspace{1mm}\hrule\vspace{1mm}
    \State //\textbf{Fine Refinement (NLL Loss)}
    \Function{NLL\_Loss}{$z, \Phi_{\text{sig}}, \{M_q\}_{q=1}^{Q-1}$}
        \State $\ell_{\text{NLL}} \leftarrow 0$
        \For{$q = 1 \dots Q-1$}
            \State $\mathcal{D}_q \leftarrow \textsc{SolveStatDist}(z, \Phi_{\text{sig}}, q)$ 
            \State $\ell_{\text{NLL}} \leftarrow \ell_{\text{NLL}} - \log(\mathcal{D}_q[M_q])$
        \EndFor
        \State \Return $\ell_{\text{NLL}}$
    \EndFunction
    \State $(\widehat{z}, \widehat{\Phi}_{\text{sig}}) \leftarrow \text{NelderMead}(\text{NLL\_Loss}, z_1, \Phi_{\text{sig},1}, \text{tol}=10^{-4})$
    
    \Statex \vspace{1mm}\hrule\vspace{1mm}
    \State \Return $(\widehat{z}, \widehat{\Phi}_{\text{sig}})$
\EndFunction
\end{algorithmic}
\end{algorithm}

\smallskip
\noindent
\textbf{Choice of Optimization Algorithm:}
We use a gradient-free Nelder-Mead simplex method \cite{nocedal2006numerical} to solve this optimization problem. Standard gradient-based solvers are inapplicable here because our numerical implementation for computing the Markov chain stationary distributions is not differentiable. Although our search space is restricted to just two dimensions $(z, \phisig)$, a brute-force grid search is computationally prohibitive; even a modest $100 \times 100$ grid would require $10,000$ evaluations of the forward model per pixel. By pairing the Nelder-Mead method with our dual loss strategy, we circumvent this bottleneck as the optimization converges in $\approx$100 iterations (detailed plots in \ref{supp:note3}). Operating in a strictly two-dimensional regime where Nelder-Mead is mathematically more robust, the algorithm drives rapid error reduction during the initial proxy-loss iterations and converges reliably during the NLL fine-tuning across varying background noise levels.
Pseudocode for our analysis-by-synthesis reconstruction algorithm is shown in Algorithm~\ref{alg:abs_pseudocode}.
Extended performance analysis in \ref{supp:note3} (Suppl. Fig.~\ref{fig:convergence_2x2}) demonstrates the necessity of this dual loss strategy; compared to an NLL-only baseline, our combined strategy prevents catastrophic depth errors at low signal-to-background ratios ($\text{SBR} \le 2$) and converges in roughly 100 iterations.

\subsection{Analysis-by-synthesis for a Pixel Array}\label{subsec:abs_pixel_array}

While the proposed optimization can be performed for each pixel individually, we can further improve results for complete scenes captured by an array of SPC pixels by using information from neighboring pixels. 
The neighboring SPC pixels capture scene points that are spatially close to each other. 
We apply a 3$\,\times\,$3 median filter regularization to improve the initial guess of the distance estimates. 
Finally, we apply a 3$\,\times\,$3 median filter to our final distance maps at the end of the iterative optimization routine.
The analysis-by-synthesis pipeline to process an array of pixels is shown in Fig.~\ref{fig:abs-pipeline}.
(See \ref{supp:note3} for details).

\begin{figure}[!t]
    \centering
    \includegraphics[width=\linewidth]{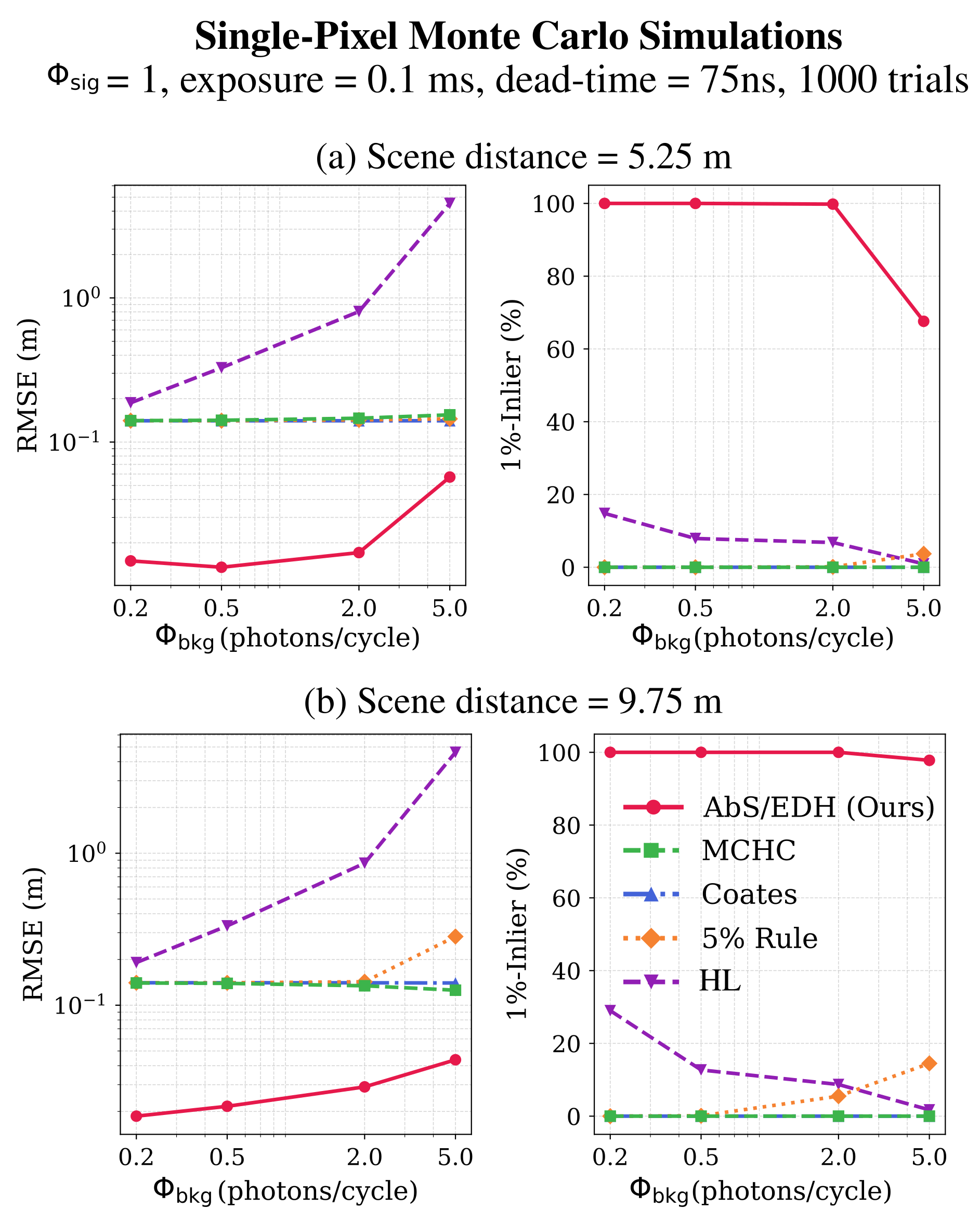}
    \caption{\textbf{Single-pixel simulation results.} The single-pixel Monte Carlo simulation results, over a varying combinations of \phisig and \phibkg,  demonstrate that our 32-bin EDH with dead-time compensation (AbS) method achieves significantly lower distance RMSE (left) and higher 1\% inliers as compared to other baselines. The MCHC \cite{rapp2021high}, Coates's correction \cite{coates1968correction}, and 5\% rule methods use 32-bin EWH measurements and are thus limited by the quantization effects. The histogramless method (HL) performs worse as the background flux increases due to nonlinearity in the measurements. }
    \vspace{-10px}
    \label{fig:multirun-abs-results}
\end{figure}

\section{Results}\label{sec:results}
In this section, we evaluate the performance and robustness of our AbS pipeline in estimating scene distance from dead-time-affected EDH measurements under different scene distances and illumination conditions. Our evaluation is structured to answer the following core questions:

\begin{itemize}
    \item[\textbf{Q1:}] How does our AbS method compare to existing baseline dead-time compensation approaches under resource-constrained settings? (Section.~\ref{subsec:monte_carlo})
    \item[\textbf{Q2:}] How sensitive is our pipeline to inaccuracies in assumed calibrated parameters? (Section.~\ref{subsec:robustness})
    \item[\textbf{Q3:}] Can our method handle physical non-idealities, such as non-Gaussian laser pulses from real hardware? (Section.~\ref{subsec:hardware})
    \item[\textbf{Q4:}] How does our approach perform in full 3D scenes containing spatial variations and multi-path interference? (Section.~\ref{subsec:multipath})
\end{itemize}

\smallskip
\noindent
\textbf{Baseline Methods:} We compare our approach against distance estimates from EWH measurements compensated using (i) Markov chain histogram correction MCHC \cite{rapp2021high}, (ii) Coates's correction \cite{coates1968correction}, and (iii) optical attenuation to reduce the photon flux as per the 5\% rule, and (iv) distance estimates obtained using histogramless lidar (HL) \cite{tontini2023histogram}.

\smallskip
\noindent
\textbf{Experiment Settings:} For all experiments, our AbS method uses EDH measurements, acquired with proportional stepping binners as explained in Section~\ref{subsec:edhspcs}, whereas the MCHC method \cite{rapp2021high} uses EWH measurements, captured in free-running mode (Section~\ref{sec:prelims}). 
The other three methods use measurements captured in synchronous mode: Coates's correction and the 5\% rule operate on EWH measurements, and the HL has access to raw timestamp measurements but does not store them. 
To the best of our knowledge, all the baseline compensation methods were originally proposed for SPCs using high-resolution EWH measurements ($\approx$ 1000 to 2000-bin EWH). 
Their performance under resource-constrained settings has never been studied. 
Our baselines must also operate under resource-constrained settings for fair comparison. 
Thus, our experiments demonstrate how these traditional estimators perform in resource-constrained settings, such as with limited laser cycles and low-resolution EWH.
(See \ref{supp:note5} for results obtained using MCHC, Coates's correction, optical attenuation using high-resolution EWH measurements without any resource limitations, CSPH\cite{gutierrez2022compressive}, and DeePEDH \cite{pedh2024}).

Considering resource estimates for EDH based on \cite{countfree2023}, both an EDH and an EWH will require about the same circuit complexity---a 32-bin EWH requires 32 registers to store histogram photon counts, while our 32-bin EDH will need 31 registers to store histogram bin boundaries.
Both approaches will use the same TDC resources. Thus, we keep the number of EDH and EWH bins equal to ensure \textit{storage-and-bandwidth-equivalent comparison}.

\begin{figure}[!t]
    \centering
    \includegraphics[width=\linewidth]{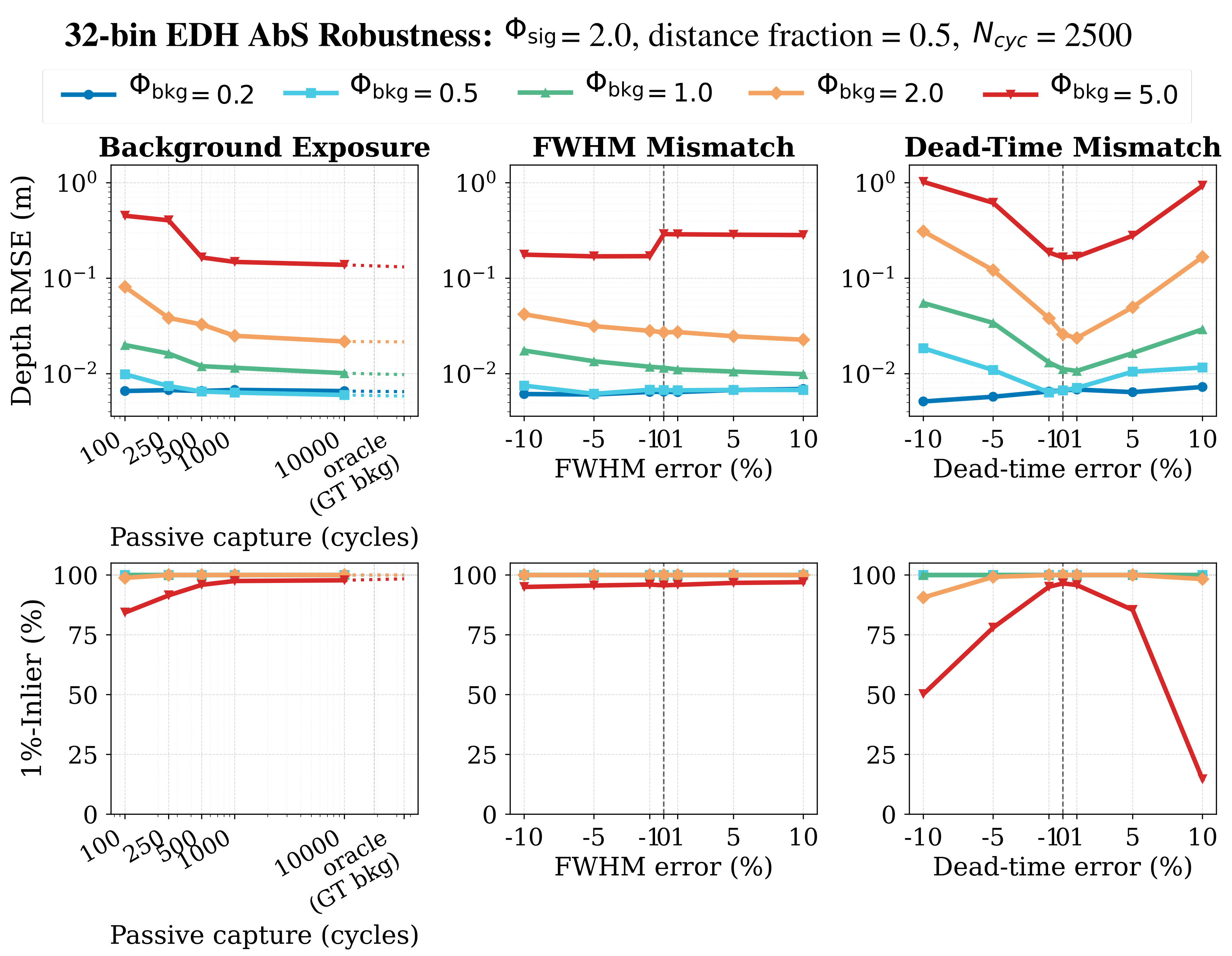}
    \caption{\textbf{Robustness of our AbS pipeline to calibration errors and model mismatch.} The AbS pipeline estimates the scene distance and signal flux by assuming known calibrated values of \phibkg, laser pulse FWHM, and $\tau_d$. 
    These simulations demonstrate a gradual increase in distance RMSE (top row) and a slight decrease in 1\% inliers (bottom row) with increasing calibration errors in the background estimate (left column), pulse FWHM (middle column), and $\tau_d$ (right column).
    The scene distance was fixed at the half the maximum distance range.f
    The smooth trends across five different illumination conditions demonstrate that our AbS estimator continues to work even in the presence of slight calibration mismatches.}
    \label{fig:robustness}
\end{figure}

\begin{figure*}[!t]
    \centering
    \includegraphics[width=1\linewidth]{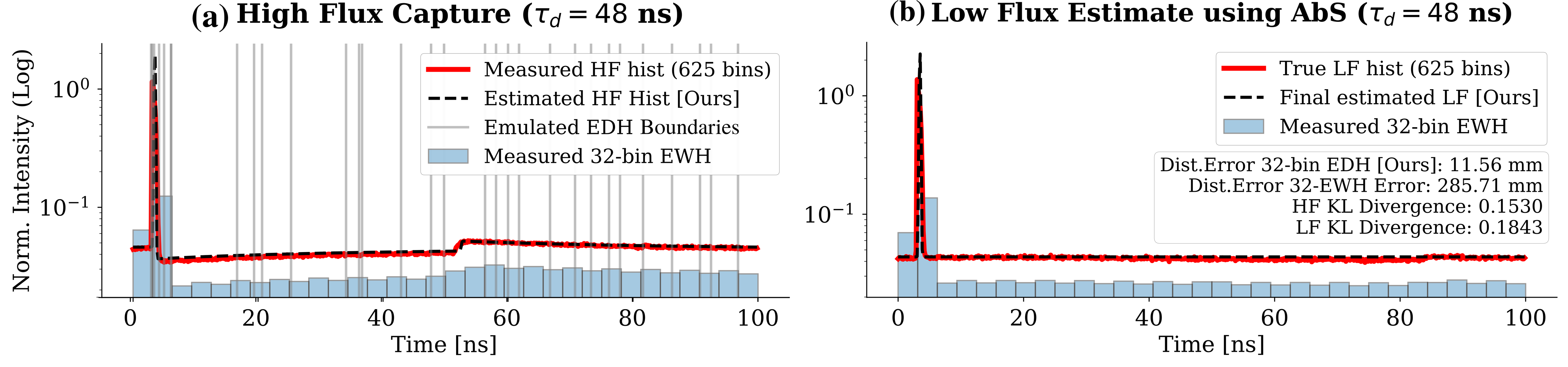}
    \caption{\textbf{AbS robustness to non-ideal transients in real hardware data.} Our AbS pipeline can handle non-ideal transients such as single-pixel transients from real hardware data. Observe that given the emulated EDH boundaries (vertical lines), there is good agreement between the actual perceived transient and the estimate in both (a) high flux (HF) conditions and (b) low flux (LF).
    In high-flux conditions, our method reliably predicts the within-peak pile-up and also the ``peak shadow'' (the dip in the transient's shape) that immediately follows the peak location, and spans a variable number of time locations depending on the dead-time.}
    \label{fig:non-ideal-transient-results}
\end{figure*}

\smallskip
\subsection{Comparisons in Controlled Simulations} \label{subsec:monte_carlo}

To answer \textbf{Q1}, we use the single-bounce imaging model described in Section~\ref{sec:deadtime_edh} to simulate single-photon timestamps, 32-bin EDH, and 32-bin EWH measurements under different combinations of known scene illumination, distance, and SPC parameters. This experiment allows us to compare the performance of our AbS pipeline with other baseline distance estimators under controlled conditions. It avoids unpredictable behaviors from non-idealities such as multi-path interference or an uncalibrated laser pulse. We use our AbS pipeline (Sec.~\ref{subsec:pixel_abs}) on the simulated EDH measurements to calculate dead-time-compensated distance estimates and compare against baseline methods and the ground truth.

Fig.~\ref{fig:multirun-abs-results} compares single-pixel results for varying combinations of \phisig, \phibkg, and distance $z_0$.
For each combination, distance estimates from all five methods were obtained using single-photon measurements simulated over 1000 independent Monte Carlo runs. The SPC measurements for each run were obtained with an exposure time of 2500 laser cycles with laser time period $T_r = 100$\,ns, Gaussian pulse with FWHM = 0.25\,ns, time locations $B = 256$, dead-time $\tau_d = 75$\,ns, scene distances $z_0 \in \{5.25, 9.75\}$\,m, $\phisig = 1.0$ and four different background flux conditions $\phibkg \in \{ 0.2, 0.5, 2.0, 5.0\}$.

The HL estimator \cite{tontini2023histogram} errors increase significantly as the nonlinearity increases with background flux. The performance of MCHC \cite{rapp2021high}, the 5\% rule, and Coates's estimator is bottlenecked by the quantized 32-bin EWH measurements. In contrast to the baseline methods, our method achieves significantly lower RMSE and a higher number of 1\% inliers.

\smallskip
\noindent
\textbf{Effect of $T_r$ on Performance:}  An EDH adaptively allocates more bins around the true peak, hence, for a fixed repetition period ($T_r$), the quantization error in estimated distance of an EDH is bounded above by that of an EWH with the same number of bins.
Results of a simulation study with varying $T_r$ are shown in \ref{supp:note4}.
EDH with our AbS pipeline maintains high distance accuracy over a wide range of $T_r$ from 2--100~ns.

In summary for \textbf{Q1}, the EWH-based SPCs suffer severe information loss when operating under resource-constrained conditions, making accurate scene distance recovery more challenging. In contrast, EDH-SPC measurements retain a significant amount of temporal information under similar constraints and dead-time distortion, enabling our AbS pipeline to recover accurate scene distances. 
We also demonstrate that our method performs better over a wide range of laser time period values and is not limited to $T_r = 100$\ ns settings. 

\subsection{Robustness to Calibration Errors}\label{subsec:robustness}

Our AbS pipeline estimates the scene distance and signal flux by assuming known, calibrated values for intrinsic and extrinsic parameters: dead-time $\tau_d$, laser pulse FWHM, and background flux \phibkg. However, Real-world measurements rarely have perfectly calibrated parameters. To address \textbf{Q2} and determine how sensitive our AbS pipeline is to calibration inaccuracies, we systematically introduced errors into these three fixed parameters.

\smallskip
\noindent
\textbf{Sensitivity to Background Flux Estimate:} The background-flux estimate is an extrinsic parameter that is scene dependent. We use a short-exposure passive capture to obtain the sensor background flux using the PF SPAD flux estimator \cite{pfspad}. The results in Fig.~\ref{fig:robustness} (first column) demonstrate that even if the \phibkg estimates are noisy due to short passive captures, the drop in performance is gradual and the performance converges rapidly towards the ideal \phibkg estimate starting from around 250 to 500 cycles (approx. 0.5 $\mu$s).

\smallskip
\noindent
\textbf{Sensitivity to Laser FWHM:} The second column in Fig.~\ref{fig:robustness} demonstrates a gradual increase in the distance RMSE when the pulse is assumed to be narrower than the true value, whereas the performance is almost similar if the pulse is assumed to be broader than the true value. These minor variations in distance RMSE and a negligible change in 1\% inliers indicate that our AbS estimator can perform well even if the laser pulse is not perfectly calibrated. 

\smallskip
\noindent
\textbf{Sensitivity to Dead-time $\tau_d$:} The third column in Fig.~\ref{fig:robustness} indicates that, unlike the other two parameters, the AbS pipeline is more sensitive to dead-time mismatches. 
However, it is important to note that, unlike the other two parameters, dead-time estimates are usually more accurate and do not deviate much from the calibrated values over time. 

In summary for \textbf{Q2}, our simulations show that the AbS pipeline degrades gracefully rather than breaking abruptly, proving it can operate effectively even in the presence of slight calibration mismatches. (See \ref{supp:note6} for additional results).

\begin{table}[!tpb]
\centering
\begin{tabular}{@{}ccccc@{}}
\toprule
\makecell{\textbf{Dead} \\ \textbf{Time}} & \multicolumn{2}{c}{\textbf{Distance Error (mm)}} & \multicolumn{2}{c}{\textbf{KL Div.}} \\ \cmidrule(lr){2-3} \cmidrule(l){4-5} 
$\tau_d$ \textbf{(ns)} & \makecell{\textbf{32-bin} \\ \textbf{EWH}} & \makecell{\textbf{32-bin EDH} \\ \textbf{[Ours]}} & \textbf{HF} & \textbf{LF} \\ \midrule
21 & 285 & 4  & 0.21 & 0.23 \\
48 & 285 & 12 & 0.15 & 0.18 \\
81 & 285 & 13 & 0.19 & 0.21 \\
89 & 285 & 8  & 0.14 & 0.15 \\
98 & 285 & 9  & 0.17 & 0.17 \\
99 & 285 & 2  & 0.23 & 0.21 \\ \bottomrule
\end{tabular}
\vspace{0.05in}
\caption{\protect\parbox[t]{\linewidth}
{\textbf{Quantitative results on real hardware.} Here we show quantitative comparison of distance error vs. ground truth distance and KL divergence of the recovered transient from the true transient across different sensor dead-times ($\tau_d$) using real hardware data from Rapp, et al. \cite{rapp2021high}.
(HF=high flux, LF=low flux)}}
\label{tab:hardware_emulation_results}
\end{table}

\begin{figure*}[!th]
    \centering
    \includegraphics[width=1.0\linewidth]{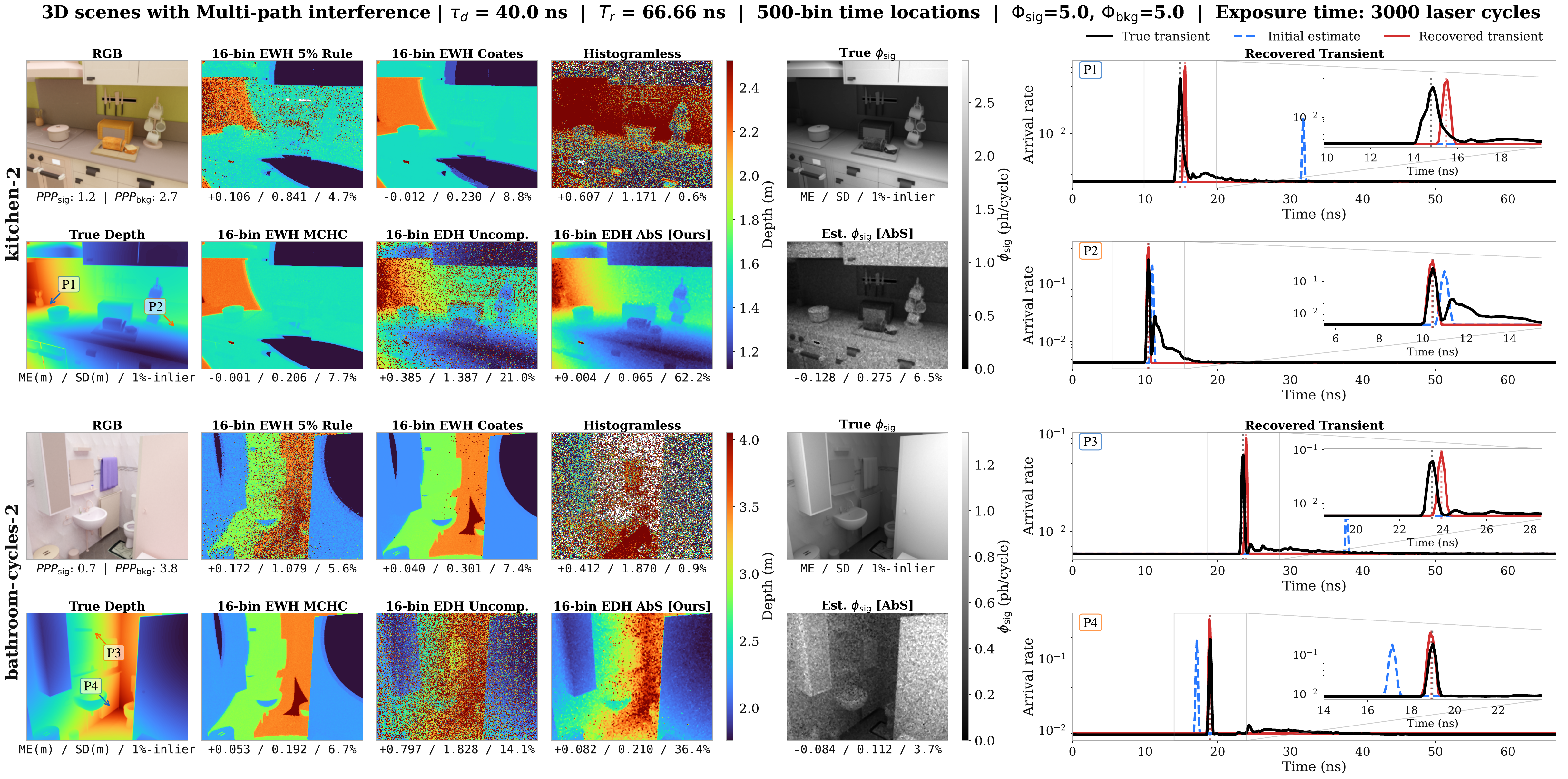}
    \caption{\textbf{Results on 3D scenes with multi-path interference from iToF2dToF dataset.} We compare our EDH-based AbS method with standard baselines that use Equi Width Histograms (EWH) and a histogramless approach. Considering resource-constrained scenarios, we keep the number of EWH bins and EDH bins equal for comparison. Quantitative results: mean error (ME, in meters), standard deviation (SD, in meters), and 1\% inlier rate (1\%in)—are reported beneath each result. As shown, our proposed method significantly reduces distance artifacts and outperforms the baselines. Additionally, we show recovered transients for some selected pixels from each scene, demonstrating how, despite the presence of multi-path interference, our method is able to converge closer to the largest peak in most scenarios.}
    \label{fig:fullscene-itof2dtof}
\end{figure*}

\subsection{Hardware Emulation Results}\label{subsec:hardware}
While the previous subsection demonstrated robustness to calibration errors and model parameter estimates, it used synthetically generated transients that assume a Gaussian pulse and a single-bounce scenario. However, real hardware measurements include physical non-idealities—most notably, non-Gaussian laser pulses. To answer \textbf{Q3}, we evaluate our AbS pipeline using a publicly available high-flux single-photon LiDAR (SPL) hardware dataset \cite{rapp2021high}, which consists of long-exposure captures of raw photon streams using a single SPAD pixel in free-running mode.

We emulate the EDH measurements using the raw photon timestamps from the single-pixel hardware captures. 
We test our single-pixel AbS pipeline on these emulated EDH measurements to estimate signal and distance.
Plugging these estimates back into our forward model, we generate an estimate of the perceived transient and compare that with the high-flux measurements (Fig.~\ref{fig:non-ideal-transient-results}(a)). 
Moreover, we also estimate the true transient (arrival rates) and compare it with the low-flux measurements (Fig.~\ref{fig:non-ideal-transient-results}(b)). 
Quantitative results for different hardware captures are shown in Table~\ref{tab:hardware_emulation_results}. 
Observe that our method recovers scene point distance with lower error than an EW-histogram with the same number of bins, and also provides transient shape recovery with low KL divergence with respect to the ground truth.
\ref{supp:note7} shows additional qualitative results and the details of the hardware emulation experiment.

To address \textbf{Q3}, we demonstrate that our AbS estimator successfully adapts to real, non-Gaussian laser pulses, achieving accurate distance recovery (errors frequently under 10 mm) compared to conventional 32-bin EWH emulated from low-flux measurements (errors exceeding 285 mm) across a wide range of dead-times.

\subsection{Transients with Multi-path Interference}\label{subsec:multipath}
While the hardware emulation results validate our method against real-world laser pulses, these measurements were captured under a controlled, single-pixel setup without any multi-path interference. Finally, to address \textbf{Q4}, we test our pipeline against spatially varying multi-path scenarios. We evaluated our method on 3D scenes from the iToF2dToF dataset\cite{itof2dtof}.
This dataset provides rendered transients obtained from a physics-based renderer (Mitsuba) \cite{NimierDavidVicini2019Mitsuba2, Pediredla2019Ellipsoidal}, resulting in complex, multi-path interference scenarios from glossy objects, interreflections, and scene regions with corners or edges. 
As shown in Fig.~\ref{fig:fullscene-itof2dtof}, our AbS pipeline, despite using a simplified analytical model, can still predict an accurate distance estimate, despite multi-path interference, with a small drop in performance compared to ideal synthetic datasets. However, the spatial distance estimates obtained using our AbS pipeline are significantly more accurate than standard uncompensated estimates as well as other baselines when operating in resource-constrained conditions.

Fig.~\ref{fig:fullscene-itof2dtof} shows qualitative and quantitative comparisons on four different 3D scenes from the iToF2dToF dataset, providing challenging scenarios with varying albedos, multipath reflections, and complex geometries. Results for more scenes and for high-resolution EWH methods are shared in \ref{supp:note5} with samples of recovered transients for selected pixels per scene. 

The results demonstrate that our proposed AbS pipeline significantly outperforms all the baseline methods. The Histogramless method completely collapses under these flux conditions, resulting in noisy distance estimates and near-zero 1\% inlier rates. The 5\% rule reduces pile-up but discards a significant amount of photons and suffers from quantization due to  16-bin EWH, resulting in highly noisy distance maps and poor structural coherence.

The Coates's estimator and MCHC successfully recover the distance estimates. However, they are also affected by the heavy quantization due to 16-bin EWH measurements, which bottlenecks their estimates. The quantization artifacts, visible as flat, discrete distance planes, limit their 1\% inlier rates to consistently below 20\%.
In contrast, our AbS pipeline effectively overcomes these severe quantization limits when operating under similar resource-constrained settings. Our method recovers smooth, continuous distance estimates and preserves geometric properties of the scene. 

In conclusion for \textbf{Q4}, despite relying on a simplified analytical model, our AbS pipeline is capable of consistently estimating accurate scene distances and is robust to the effect of multi-path interference in 3D scenes. It achieves the highest 1\% inlier rates across all scenes (reaching up to 62\% in the Kitchen scene), demonstrating its effectiveness in complex light transport scenarios when operating under pixel memory constraints.

\section{Discussion and Future Work}\label{sec:conclusion}

\smallskip
\noindent\textbf{Convergence to Steady-state:} The theoretical results in \ref{supp:theory} demonstrate that the EDH control values obey a Markov transition matrix, hence, provide an upper bound on the convergence rate of the EDH CVs (the convergence is $O(|\lambda_2|^n)$ where $\lambda_2$ is the second largest eigenvalue of the tri-diagonal transition matrix and $n$ is the number of photons detected) \cite{levin2026markov}. 
Empirical simulation results in \ref{supp:theory} suggest that the EDH converges in 500--2500 laser cycles over a wide range of flux levels of interest.
Assuming a $15\;\text{MHz}$ laser repetition rate ($10$~meter range), the exposure time is $\approx$ 50--250 microseconds depending on the total flux. 
Scene motion and illumination changes can be safely ignored at such timescales.

\noindent\textbf{Computational Complexity and Runtime Estimates:} The basic building block of EDH is an extremely lightweight in-pixel computation where the control value is updated (incremented/decremented) on each photon detection.
A key conceptual novelty is that the EDH representation contains sufficient information for pile-up compensation.
However, the proposed compensation algorithm is not intended to run in-pixel.
Though a direct application of
AbS
that includes recomputations of the stationary distributions of the binner CVs at each evaluation of the negative log-likelihood has a high cost
($O(QB^3)$ where $Q$ is the number of quantiles and $B$ is the size of the discrete time grid), we show in \ref{supp:loss} that other cost functions have a cheaper $O(Q)$ cost.
The AbS pipeline developed here shows that ED histograms contain sufficient information to enable reliable estimation of distance and scene reflectivity even in the presence of dead-time distortions.
In the future, we envision the pile-up compensation could be done using non-iterative approximations to the full AbS pipeline (e.g., a feed-forward DNN).
Moreover, the success of a simple 3$\times$3 spatial median filter indicates that stronger spatial priors learned by a DNN can further improve accuracy. 

\noindent\textbf{Towards Resource-efficient and Robust SPCs}: We tackle the challenges of dead-time-induced pile-up distortion and compression jointly by proposing a hardware-algorithm co-design approach that runs SPC pixels in a free-running mode and estimates scene properties (distance and reflectance) using an analysis-by-synthesis pipeline.
We demonstrate that compressed ED histogram representations contain sufficient information to compensate for dead-time distortions.
Although we do not make any claims about our analysis-by-synthesis algorithm's computational efficiency, our results demonstrate that at least one type of compressive histogram contains sufficient information to accurately recover scene distance.
It will be worthwhile to investigate more optimized approaches that can run in or close to the SPC sensor pixels.
Although we chose ED histogram as the compressed representation in this work, it will be worth developing pile-up compensation techniques for other approaches such as linear projections \cite{gutierrez2022compressive} and sketching \cite{sheehan2024spline}.

\ifpeerreview \else
\section*{Acknowledgments}
This work was supported in part by NSF ECCS 2138471 and the Portland State University Venture Development Fund.
We thank Keylan Petty for assistance with initial exploratory simulations on the effect of dead-time on ED histogrammers.
\fi

\bibliographystyle{IEEEtran}
\bibliography{references}

\ifpeerreview \else

\fi

\clearpage
\onecolumn
\renewcommand{\figurename}{Supplementary Figure}
\renewcommand{\thesection}{Supplementary Note \arabic{section}}
\renewcommand{\theequation}{S\arabic{equation}}
\setcounter{figure}{0}
\setcounter{section}{0}
\setcounter{subsection}{0}
\setcounter{equation}{0}
\setcounter{page}{1}

\begin{center}
\Large Supplementary Document for\\
\Large ``High-Flux Count-Free Single-Photon 3D Cameras'' \\[0.2cm]
\large Kaustubh Sadekar$^a$, Vivek K Goyal$^b$, David Maier$^a$, Atul Ingle$^a$\\[0.1cm]
\large  $^a$Portland State University $\;\;\;\;\;\;\;\;^b$Boston University\\[0.1cm]
\normalsize Email: \texttt{\{ksadekar, maier, ingle2\}@pdx.edu}; $\;\;\;$ \texttt{goyal@bu.edu}
\end{center}

\section{Binner-Control-Value Stationary Distribution: Theory and Simulations}\label{supp:theory}

The basic building block of the equi-depth histogrammer's (EDH) compressive capture method is a binner circuit that tracks arbitrary quantile positions of the underlying perceived transient.
The binner circuit's control value (CV) is updated with each photon detection and can be modeled as a Markov chain, as discussed in the main text.
The mode of the stationary distribution of a binner's CV Markov chain is at the actual quantile position of the underlying perceived transient.
In the following, we present a theoretical proof for the median-tracking binner.
The result generalizes easily to other quantiles as well.
Monte Carlo simulation results show good agreement between the actual quantiles of the perceived transient and the modes of the corresponding binner CV stationary distributions for 31 different quantiles of a 32-bin EDH.

\subsection{Theoretical Results}

We prove a theoretical result showing that the stationary distribution of the control value of a median-tracking binner is monotonically increasing, then monotonically decreasing, with a global maximum at the actual quantile position of the perceived transient $\widetilde{\Phi}$.
For simplicity, we present a proof below for the median-tracking binner.
The same proof can be easily extended to any arbitrary quantile that is different from the median.

Let $\Lambda_j = \sum_{i=1}^j \widetilde{\Phi}[i]$. 
Since the perceived transient $\widetilde\Phi$ is itself the stationary distribution of the photon detection Markov chain, $\Lambda_0 = 0$, $\Lambda_B = 1$, and $\Lambda_j$ is monotonically increasing $\Lambda_j < \Lambda_{j-1}$.

We make the following assumption to ensure that there is indeed a discrete index corresponding to the actual median of the perceived transient.

\smallskip
\noindent\textbf{Assumption:} $\exists$ $1 < j^* < B$ such that $\Lambda_{j^*} = 1-\Lambda_{j^*} = 1/2$.

Note that this assumption is not too limiting in practice. Since the underlying photon detection process is a continuous-time Markov chain [Suppl. Ref. \ref{ref_rapp}], we can create a discretization on a finer grid (by increasing $B$) to ensure that a $j^{*}$ exists such that $\Lambda_{j^*}$ is arbitrarily close to $1/2$.

\smallskip
\noindent\textbf{Theorem 1:} The Markov chain of the median-tracking binner's control value is irreducible, aperiodic, and reversible.

\noindent
\textbf{Proof:}
The CV Markov chain is irreducible because any node can be reached from any other node; the graph is strongly connected.
The chain is aperiodic because it has self-loops in each state with non-zero probabilities of staying in the same state.
It follows that the Markov chain must have a stationary distribution [Suppl. Ref.~\ref{ref_grimmett}]  which we denote by $\mathcal{D}$ (following the notation in the main text, and dropping the $q=1/2$ subscript for convenience). 
We now prove that the chain is reversible, i.e., it obeys the detailed balance equation:
$\mathcal{D}[i] p_{i,i+1} = \mathcal{D}[i+1] p_{i+1,i}$.
Since $\mathcal{D}$ is the stationary distribution, we have:
$$
\mathcal{D}[i] = \sum_j \mathcal{D}[j] p_{j,i}.
$$
Since each node in this chain is only connected to its immediate neighbors and to itself via a self-loop, we can simplify this to:
$$
\mathcal{D}[i] = \mathcal{D}[i-1] p_{i-1,i} + \mathcal{D}[i+1] p_{i+1,i} + \mathcal{D}[i] p_{i,i}.
$$
Rearranging terms, we get:
$$
\mathcal{D}[i] (1-p_{i,i}) = \mathcal{D}[i-1] p_{i-1,i} + \mathcal{D}[i+1] p_{i+1,i}.
$$
Since at any node $i$ the probabilities of staying at the node and the probabilities of leaving the node sum to 1, we have $p_{i,i} + p_{i,i-1} + p_{i,i+1} = 1$, which implies $1-p_{i,i} = p_{i,i-1} + p_{i,i+1}$.  Therefore,
$$
\mathcal{D}[i] (p_{i,i-1} + p_{i,i+1}) = \mathcal{D}[i-1] p_{i-1,i} + \mathcal{D}[i+1] p_{i+1,i}.
$$
Plugging in $i=0$, we get:
$$
\mathcal{D}[0] p_{0,1} = \mathcal{D}[1]p_{1,0}
$$
showing that detail balance holds for state $0$.
Next, plugging in $i=1$, we show that detail balance holds for state $1$ as well:
\begin{align*}
\mathcal{D}[1] (p_{1,0}+p_{1,2}) &= \mathcal{D}[0] p_{0,1} + \mathcal{D}[2] p_{2,1} \\
   \Rightarrow \mathcal{D}[1] p_{1,0}+ \mathcal{D}[1] p_{1,2} &= \mathcal{D}[0] p_{0,1} + \mathcal{D}[2] p_{2,1} \\
   \Rightarrow \mathcal{D}[0] p_{0,1}+ \mathcal{D}[1] p_{1,2} &= \mathcal{D}[0] p_{0,1} + \mathcal{D}[2] p_{2,1} \;\;\;\;\;\;\;\; (\dagger)\\
   \Rightarrow  \mathcal{D}[1] p_{1,2} &=  \mathcal{D}[2] p_{2,1}
\end{align*}
where $(\dagger)$ follows from detail balance for state $i=0$.
By continuing this recurrence, we can inductively show that detail balance holds, in general, for any state $i$:
$$
\mathcal{D}[i] p_{i,i+1} = \mathcal{D}[i+1] p_{i+1,i}.
$$
We remark that this proof only relied on the birth-death chain structure of the CV Markov chain, and not on the exact values of the transition probability matrix, which we will exploit in the next result.

\smallskip
\noindent\textbf{Theorem 2:} For a median-tracking binner, $\mathcal{D}[i-1] < \mathcal{D}[i]$ for $i < j^*$ and $\mathcal{D}[i-1] > \mathcal{D}[i]$ for $i > j^{*}$. 

\noindent\textbf{Proof:}
From Theorem 1, since the CV Markov chain is reversible, the detail balance condition implies:
$$
    \mathcal{D}[j] = \mathcal{D}[j-1] \frac{p_{j-1,j}}{p_{j,j-1}}.
$$ 
Since the transition matrix entries are $p_{j-1,j} = \frac{1}{2}(1-\Lambda_{j-1})$ and $p_{j,j-1} = \frac{1}{2}\Lambda_j$, we get
$$
    \mathcal{D}[j] = \mathcal{D}[j-1] \frac{(1-\Lambda_{j-1})}{\Lambda_j}. \;\;\;\;\;\; (\ddagger)
$$
We now exploit the unique structure of the median-tracking binner's CV Markov chain.
The state transition probabilities depend on $\Lambda_j$'s that are themselves cumulative densities.
Observe that since $\Lambda_j$ is a cumulative sum, $\Lambda_j > \Lambda_{j-1}$ which implies $1-\Lambda_j < 1-\Lambda_{j-1}$.
When $j < j^*$, $\Lambda_j < 1-\Lambda_j$.
This implies $\Lambda_j < 1 - \Lambda_j < 1-\Lambda_{j-1}$ which implies that
$$
\frac{1-\Lambda_{j-1}}{\Lambda_j} > 1. \;\;\;\;\;\;\;\; (\S)
$$
From $(\ddagger)$ and $(\S)$ we get:
$$
D[j] > D[j-1].
$$

When $j > j^*$ we get the opposite inequality.
We start with the detail balance equation:
$$
    \mathcal{D}[j+1] = \mathcal{D}[j] \frac{(1-\Lambda_{j})}{\Lambda_{j+1}}.
$$
Since $\Lambda_j$ is monotonically increasing, $\Lambda_j < \Lambda_{j+1}$.
For $j > j^{*}$, $1-\Lambda_j < \Lambda_j$.
Therefore, $1-\Lambda_j < \Lambda_{j+1}$ which imples
$$
\frac{1-\Lambda_j}{\Lambda_{j+1}} < 1.
$$
Therefore, for $j>j^*$ we get:
$$
\mathcal{D}[j+1] < \mathcal{D}[j].
$$

It follows immediately that $\mathcal{D}[j]$ is maximum when $j=j^*$.

\smallskip
\noindent \textbf{Corollary:} The mode of the median-tracking binner's stationary distribution is $\mathcal{D}[j^*]$.

\smallskip
\noindent \textbf{Theorem 3:} Assume $\exists \; 1<j_q^*<B$ such that
$\Lambda_{j_q^*} = 1-\Lambda_{j_q^*} = q$ where $0<q<1$.
The mode of the binner that tracks the $q^\text{th}$ quantile of the perceived transient is $\mathcal{D}_q[j_q^*]$.

\noindent \textbf{Proof:} The proof is similar to the median-tracking binner's proof where we replace the $1/2$ in the transition probability terms with unequal ``weighting'' terms of $q$ and $1-q$ as shown in Fig. 5(b) of the main text.

\subsection{Empirical Simulations}

We ran single-pixel, Monte Carlo simulations of quantile-tracking binners over different combinations of scene distance, signal strengths, background strengths, and dead-times, and compared the empirical distributions of the final CVs with those obtained from our analytical model of the Markov chain stationary distributions $\mathcal{D}_q$.

\begin{figure}[!ht]
    \centering
    \includegraphics[width=0.8\linewidth]{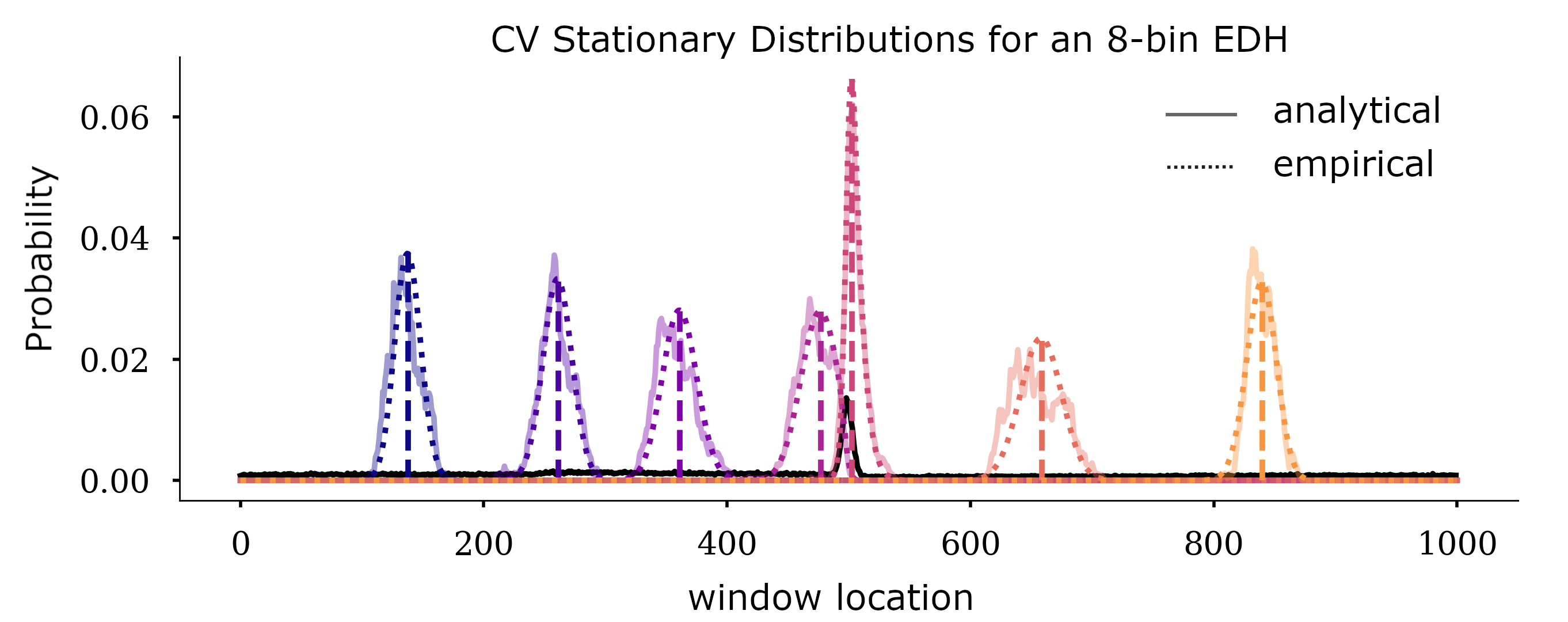}
    \caption{\textbf{Plots showing empirical histograms of final CV boundaries over multiple runs vs analytical CV stationary distributions.}
    The x-axis in this plot denotes the discretized window locations (length 1000) and the y-axis denotes the probability that the CV is found at a given window location.
    There is good agreement between the analytical model (dotted lines) and the empirical Monte Carlo simulations of the binner CV (solid jagged lines).
    The vertical dashed lines denote the true quantile positions, which line up with the CV stationary distributions' modes.}
    \label{fig:emp-vs-an}
\end{figure}

We simulated 7 binners tracking the quantiles of an 8-bin EDH, for multiple independent runs, and stored the final CV locations for each of the runs.
Finally, we plot these empirical distributions of the simulated binners and the numerically computed stationary distributions of the CV locations for the corresponding quantiles.
See Suppl. Fig.~\ref{fig:emp-vs-an}.
The plot shows good agreement between the empirical and analytical stationary distributions.
Specifically, note that the modes of these distributions align quite well.

To empirically validate the theoretical result on the mode, we conducted extensive simulations across 648 different combinations of signal strength, background, dead-time, and ground-truth distance values for a 32-bin EDH.
Using $T_r = 100$ ns and number of windows locations $B = 500$,
distance was varied from 10\% to 90\% of the maximum unambiguous distance range in steps of 10\%. 
The rest of the parameters were chosen as follows: laser pulse width FWHM (nanoseconds) $\in \{1.0, 2.5, 5.0\}$, dead-time (nanoseconds) $\in \{0.0, 25, 75\}$,
$\phisig \in \{0.1, 1.0\}$, 
and SBR $\in \{2.0, 1.0, 0.5, 0.2\}$.

\begin{figure}[!ht]
    \centering
    \includegraphics[width=1\linewidth]{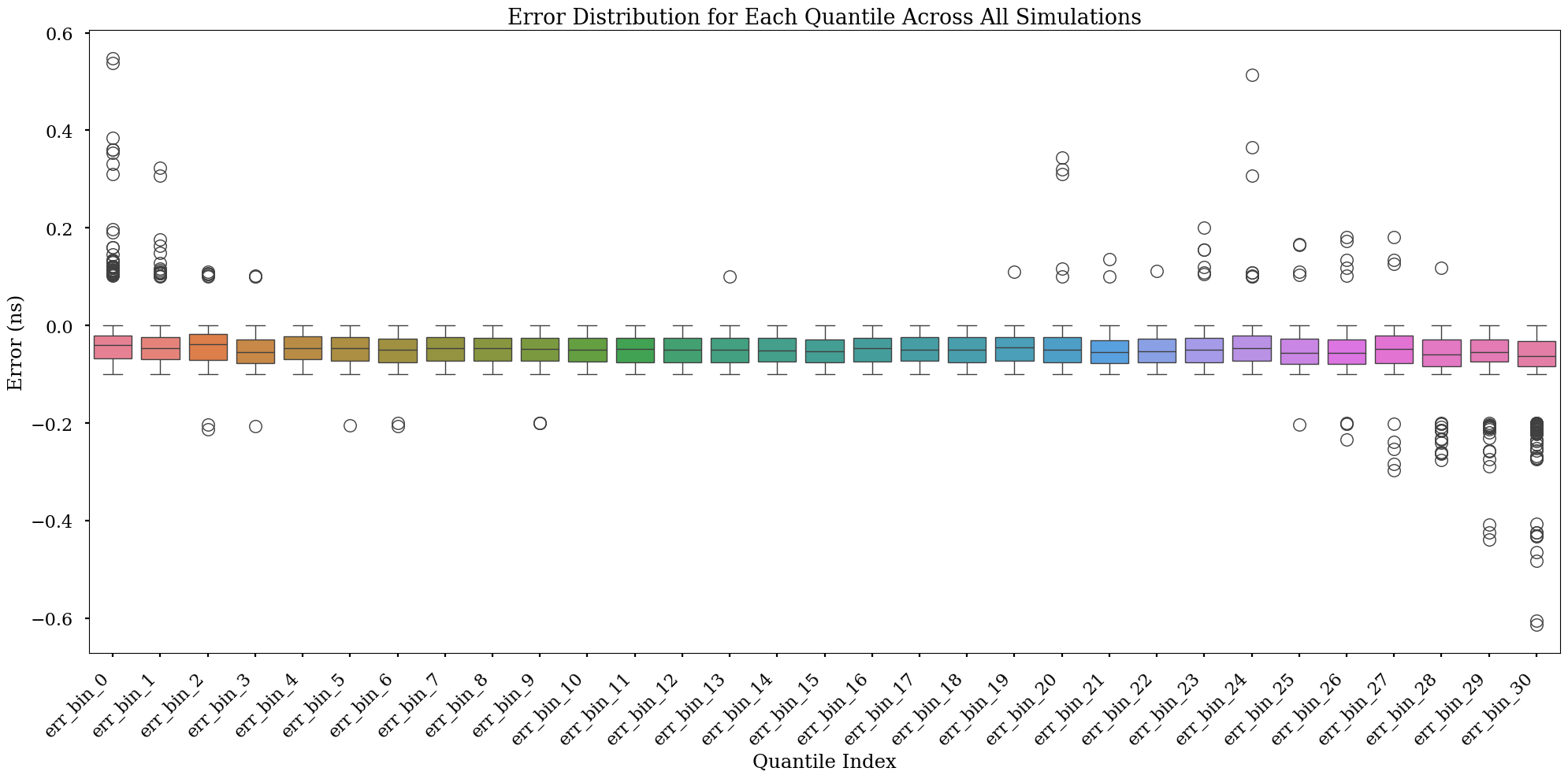}
    \caption{\textbf{Error plot showing the difference between the mode of the stationary distribution $\mathcal{D}_q$ and the $q^{th}$ quantile of the empirically simulated perceived transient, aggregated over different combinations of distance, signal strength, background strength, and dead-time.}
    We show the spread of errors for each of the 31 quantiles that form a 32-bin EDH.
    Observe that the errors are quite small (hovering around the zero line), indicating good agreement between theory and empirical simulations.}
    \label{fig:stn_dist_modes}
\end{figure}

We calculate the difference error between the modes of the analytically-derived stationary distributions of the 31 binners, and the locations of the corresponding quantiles of the perceived transients.
Observe the small error spread in Suppl. Fig.~\ref{fig:stn_dist_modes}, indicating good agreement between theory and simulation.

\clearpage

\clearpage \newpage

\section{Loss functions}\label{supp:loss}

The results shown in the main paper use the negative-log-likelihood (NLL) loss function to estimate the true scene parameters from a set of EDH measurements.
Although theoretically sound, calculating this loss function can be cumbersome in practice due to the $O(QB^3)$ cost of numerically computing the stationary distributions of each of the $Q-1$ quantile-tracking binners that form the EDH\@.
We experimented with two other loss functions that provide approximate alternatives for the NLL loss, and have a lower, $O(Q)$, cost.

\smallskip \noindent
\textbf{Boundary MSE:}
We simply calculate the mean-squared error (MSE) of boundary positions, which we call the boundary-MSE  loss defined as:
\begin{equation}\label{eq:brmse}
\mathcal{L}_\text{BMSE} =  \frac{1}{Q-1} \sum_{q = 1}^{Q-1} \| M_{q} - C_{q} \|^2,
\end{equation}
where $C_q$ are the quantiles of the perceived transient synthesized from the forward model.
Recall, from the main text, that $M_q$ denotes the measured quantile location from the EDH, and $C_q$ is the quantile location of the perceived transient.
Although intuitive and straightforward to implement, the boundary-MSE loss does not account for the fact that $M_q$ are estimates of quantiles, and a small error in the position of $M_q$ may correspond to a large error in the quantile in regions of the transient where the photon density is high.
This motivated us to calculate the MSE in terms of how far off we are in the ``quantile space,'' instead of the absolute positions.

\smallskip
\noindent \textbf{Quantile MSE}:
The intuition behind quantile MSE is that a deviation of even a small $\pm 1$ difference in boundary position can cause a much larger quantile error if the quantile is close to the laser peak where the photon density is high.
(Conversely, a relatively large deviation in boundary position might correspond to a small quantile error for boundaries away from the peak.)  
We propose quantile-MSE $\mathcal{L}_\text{QMSE}$, which computes the error in terms of the quantile position instead of the absolute boundary positions:
\begin{equation}
    \mathcal{L}_\text{QMSE} = \frac{1}{Q-1} \sum_{q = 1}^{Q-1} \Big\lVert \frac{q}{Q} -\sum^{M_{q}}_{k=0} \widetilde{\Phi}[k] \Big\rVert^2,
\end{equation}
where $\widetilde{\Phi}$ is the perceived transient (which corresponds to, say, the current guess of the unknown signal and distance values in the AbS pipeline).

\begin{figure}[!ht]
    \centering
    \includegraphics[width=0.9\linewidth]{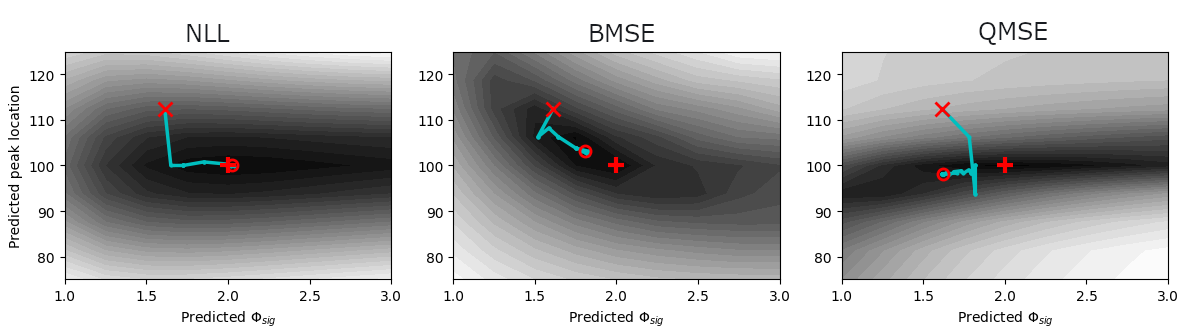}
    \caption{\textbf{Example loss-function contour maps}. We show three different loss functions (negative-log-likelihood, boundary MSE and quantile MSE) in a simulated example.
    NLL gives the best match between the final optimizer estimate and the true value but has a higher computational overhead.
    Boundary MSE provides reasonable estimates at slightly higher error.
    Quantile MSE provides reliable distance estimate but poor signal strength estimates.
    \textcolor{red}{$\times$} denotes the starting guess in the AbS pipeline, \textcolor{red}{$\circ$} denotes the final converged position, and \textcolor{red}{$+$} denotes the true signal and distance combination used in this specific example.
    The trajectory of estimates over each iteration of the optimizer is overlaid on the cost-function contours.}
    \label{fig:loss_contours}
\end{figure}

Some example loss-function contours are shown in Suppl. Fig.~\ref{fig:loss_contours}.
These were generated from a true transient with $\phisig = 2.0$, $\phibkg = 2.0$, $\tau_d = 75$\,ns, laser FWHM = $2.5$\,ns, $T_r=100$\,ns, and a 32-bin EDH\@.
In general, we observed that the NLL loss function is quite robust to noise, in that the valley of the overall loss function landscape lies quite close to the true values.
Since boundary-MSE ignores the underlying distribution of the CV stationary distributions and instead collapses it to a single number, the quantile of the perceived transient, we notice slightly larger errors in the final estimates.
We observed that the quantile-MSE loss landscape often gave skinny and long valleys parallel to the signal-strength axis.
Using the quantile-MSE loss gives larger errors in the signal-strength estimates, albeit reliable estimates of the distance.

\clearpage \newpage
\section{Analysis-by-Synthesis Pipeline: Implementation Details}\label{supp:note3}

In this section we describe implementation details of the analysis-by-synthesis (AbS) reconstruction pipeline used for full scene reconstructions shown in the last figure in the main text.

\smallskip
\noindent
\textbf{Simulation from RGBD Datasets:}
Starting with existing ground-truth datasets (like NYUv2 and Middlebury) that provide RGB and distance maps, we simulated dead-time-distorted measurements captured by an SPC under varying signal, background illumination, and dead-time conditions.
Since most commercial SPCs use an infrared laser, we use the red channel in the RGB image as an approximate proxy for the scene pixels' reflectance scaling terms, and apply distance-squared falloff to change the effective signal and background photon flux received by each pixel.
We mention a single signal strength value $\phisig$ for our full-scene simulation results.
This value indicates the signal photons seen by a (hypothetical) scene patch that is located exactly 1 meter away and has a reflectance of 1.0.
Our laser simulation parameters are set to a FWHM of 2.5 nanoseconds and a repetition period of $T_r$ of 100 nanoseconds.
We show results with different dead-time conditions ranging from 0 (no dead-time) to 100 nanoseconds.
Our simulation code will be open-sourced after paper acceptance.
The EDHs are simulated on a per-pixel basis by randomly sampling each binner's CV stationary distributions.

\smallskip
\noindent
\textbf{Parameter Initialization:} 
The AbS pipeline relies on iterative optimization to find the signal/distance combination that most closely matches the set of EDH measurements.
A good initial guess for the unknown signal and distance values helps speed up this optimizer's convergence. 
Since the EDH inverse-bin-widths roughly correspond to the underlying photon detection ``density,'' we use the midpoint of the narrowest ED bin as our initial estimate of the distance (laser peak location). 
We apply a $3 \times 3$ median filter on these narrowest-bin distance estimates over a pixel neighborhood and use it as the initial guess $z_\text{init}$ for that pixel's distance.
Next, we estimate the effective background photons $\PhiHatbkg$ using the known SPC dead-time value and the photon counts from measurements captured in a passive setting (without using the laser source). 
We then compute $\PhiHatsiginit$ the initial estimate for the effective signal using $\PhiHatbkg$ and  $z_\text{init}$.

\smallskip
\noindent
\textbf{Optimization details:} The optimization uses the measured ED bin locations from the EDH, the fixed (known) parameters of the forward model, like dead-time, laser time period, and laser pulse width, and uses $\PhiHatbkg$ as a proxy for the true background flux.
We suspect that the results might improve slightly if $\phibkg$ were to also be estimated jointly in a three-parameter optimization routine, but we decide to use the passive capture estimate because of its simplicity and practical efficacy.
We use a gradient-free Nelder-Mead optimization routine in Python's \texttt{scipy.optimize} library to estimate the scene distance $z$ and effective signal $\PhiHatsig$.

\smallskip
\noindent
\textbf{Postprocessing:}
As a final filtering stage, we apply a $3\times 3$ median filter to both the estimated signal and estimated distance maps.
We see the median filters used in the initialization and post-processing stages as a simple form of regularization applied on top of the optimization routine that maximizes the negative-log-likelihood.
Future work will explore other ways of regularization (such as total-variation loss, or using neural networks to incorporate scene-priors).

In the main text, we described our analysis-by-synthesis (AbS) pipeline for estimating scene distance $z$ and signal strength $\phisig$. Because the forward model $\mathcal{F}$ relies on numerical computation of Markov chain stationary distributions, standard gradient-based optimization is inapplicable. Here, we provide detailed performance profiling for our chosen gradient-free solver, the Nelder-Mead simplex method. 

\subsection*{Performance Analysis of the Dual Loss Optimization Strategy}

To balance computational efficiency with theoretical accuracy, we employ a dual loss optimization strategy within the AbS optimization (Stage~2 of the pipeline). 
First, we run the Nelder-Mead optimizer using the boundary-MSE loss $\mathcal{L}_\text{BMSE} $ (which has $\mathcal{O}(Q)$ complexity) to rapidly traverse the search space and locate the general global minimum. 
Once the simplex has collapsed into the correct neighborhood, we switch to the exact negative log-likelihood loss ($\mathcal{L}_\text{NLL}$) for the final fine-tuning iterations.
This dual loss optimization ensures that we only compute the expensive $\mathcal{O}(QB^3)$ stationary distributions when the estimate is near convergence.

To characterize the per-pixel convergence behavior of the AbS optimizer, we profile Nelder-Mead iteration counts and distance RMSE across a sweep of background flux conditions ($\Phi_\text{bkg} \in \{0.2, 0.5, 1.0, 2.0\}$, SBR~$10$--$1$) for two target distances: one-fourth the maximum distance range, which falls within the dead-time shadow of the $75\,\text{ns}$ dead-time, and one-half the maximum distance range (at the center of the 100 nanosecond laser period). 
Each pixel receives an EDH observation drawn from the stationary detection distribution (known background, $200$ independent pixels per experiment setting), and a $3\times3$ spatial median filter is applied to the narrowest-bin distance initialization prior to optimization.

\begin{figure}[t]
    \centering
    \includegraphics[width=\linewidth]{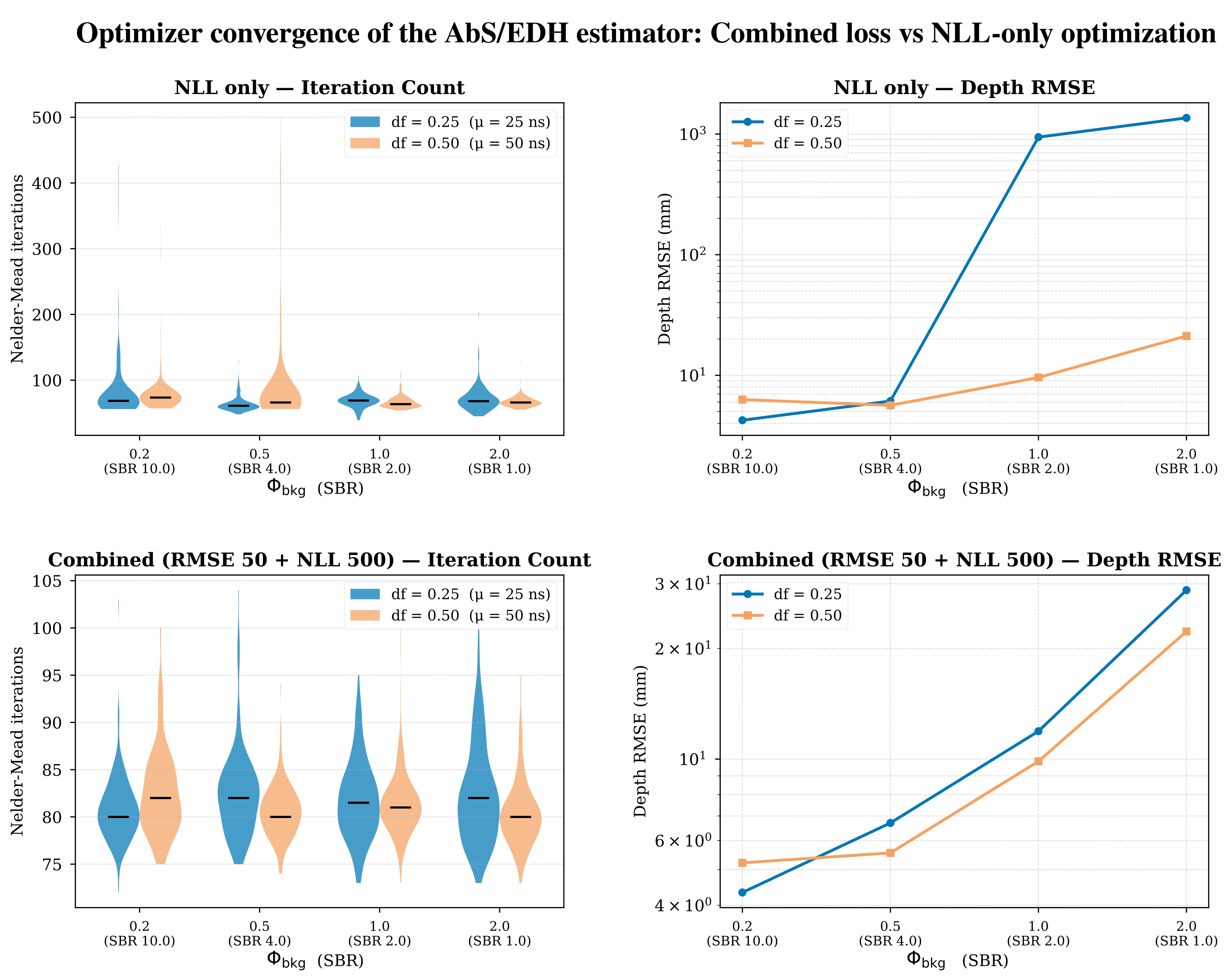}
    \caption{
        \textbf{Optimizer convergence of the AbS/EDH estimator:}
        Each panel sweeps background flux $\Phi_\text{bkg} \in \{0.2, 0.5, 1.0, 2.0\}$
        (\text{SBR}~$10$--$1$) for two target distances at one-fourth and one-half the maximum distance range, with a $3\times3$ spatial median filter applied
        to the distance initialization.
        \emph{(Left)} Nelder-Mead iteration count distributions;
        \emph{(right)} distance RMSE.
        The combined loss optimization (50-iteration RMSE followed by 500-iteration NLL)
        maintains sub-$30\,\text{mm}$ RMSE and a tight iteration spread across all
        conditions, whereas NLL-only degrades catastrophically at $\text{SBR} \leq 2$
        for the shallower target (at one-fourth the maximum distance range), where dead-time distortion
        is most severe.
        All results use $\Phi_\text{sig} = 2$, dead-time $= 75\,\text{ns}$,
        $T_\text{max} = 100\,\text{ns}$, $200$ independent pixels per condition.
    }
    \label{fig:convergence_2x2}
\end{figure}

We compare two loss configurations: NLL-only and the combined dual loss optimization ($50$-iteration $\mathcal{L}_\text{BMSE} $ optimization followed by NLL fine-tuning). 
As shown in Suppl.~Fig.~\ref{fig:convergence_2x2}, the combined dual loss achieves consistent sub-$30\,\text{mm}$ distance RMSE and a tight distribution of iterations across both distance fractions. 
In contrast, NLL-only method performs comparably at half the maximum distance range but degrades severely at one-quarter of the maximum distance range, for $\text{SBR} \leq 2$.
\clearpage \newpage
\section{Effect of Laser Time Period $T_r$}\label{supp:note4}

We evaluate how depth estimation quality scales with the laser time period
$T_r$ for EDH+AbS and three EWH-based baselines (MCHC, Coates, 5\% Rule).
We sweep $T_r \in \{2, 5, 25, 50, 100\}$\,ns at three scene distances
(7.5, 15, and 22.5\,cm) with no dead-time distortion ($\tau_d = 0$), isolating
the effect of $T_r$ from pile-up.
The laser pulse width is set to $\text{FWHM} = 0.01\,T_r$, keeping the pulse
at a constant 5-bin resolution in the forward model across all conditions.
All methods use 16 bins ($n_\text{EDH} = n_\text{EWH} = 16$); signal and
background flux are fixed at $\Phi_\text{sig} = \Phi_\text{bkg} = 2.0$
photons/cycle, and results are averaged over 100 independent pixels.

\begin{figure}[h]
    \centering
    \includegraphics[width=\linewidth]{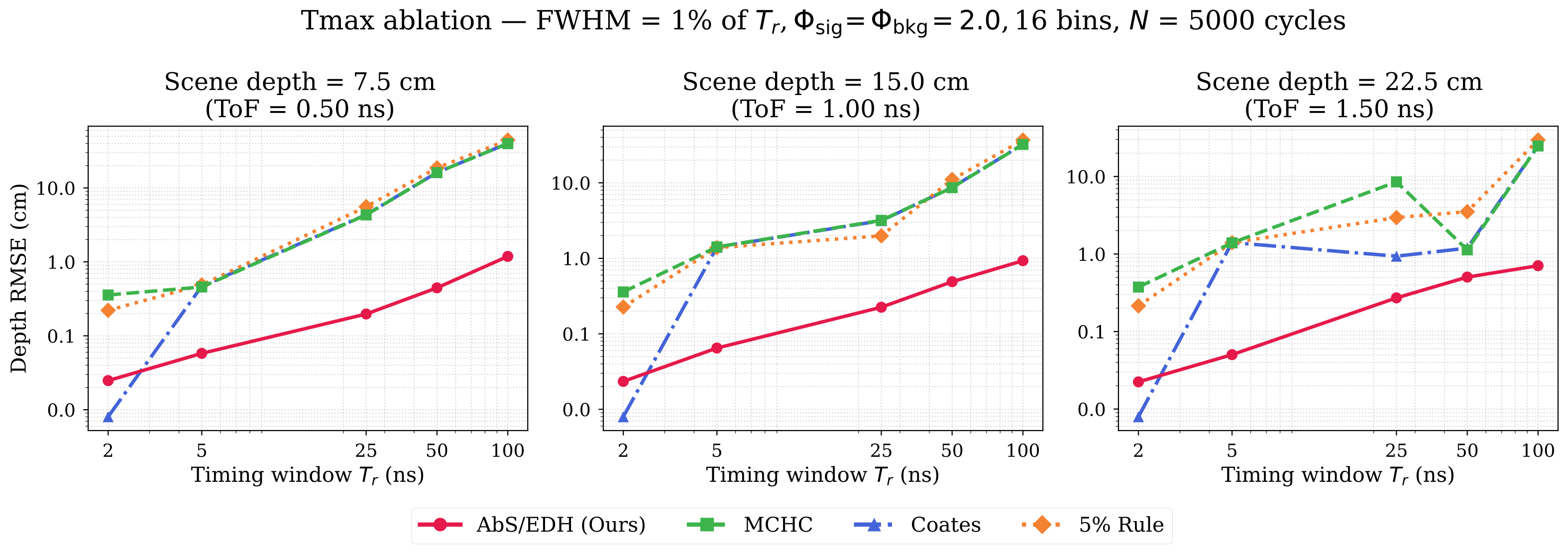}
    \caption{
        \textbf{Depth RMSE vs.\ laser time period $T_r$ for three scene
        distances (7.5, 15, 22.5\,cm) with no dead-time distortion
        ($\tau_d = 0$).}
        The laser pulse width scales with $T_r$ ($\text{FWHM} = 0.01\,T_r$)
        to maintain constant pulse resolution across all conditions.
        All methods use 16 bins; $\phi_\text{sig} = \phi_\text{bkg} = 2.0$
        photons/cycle, 100 pixels per condition.
        EDH+AbS (ours) maintains near-constant RMSE as $T_r$ grows, while
        EWH-based methods degrade proportionally as their bin width
        $T_r/n_\text{EWH}$ widens linearly with the laser time period.
        The result holds across all three scene distances.
    }
    \label{fig:supp:tmax_ablation}
\end{figure}

As shown in the Suppl. Fig.~\ref{fig:supp:tmax_ablation}, EDH+AbS maintains near-constant
RMSE as $T_r$ grows, at 15\,cm, RMSE ranges from 0.02\,cm at $T_r = 2$\,ns
to 0.93\,cm at $T_r = 100$\,ns, while EWH-based baselines degrade by one to
two orders of magnitude over the same range, reaching 32--37\,cm at
$T_r = 100$\,ns.
This gap arises because EWH bin width grows linearly with $T_r$ ($T_r/n_\text{EWH}$),
eventually spanning many times the pulse width and reducing depth readout to
coarse bin-center quantization.
EDH's quantile representation adaptively concentrates all 16 boundaries near
the signal peak regardless of $T_r$, making depth precision insensitive to the
laser time period.
The trend is consistent across all three scene distances.

\clearpage
\newpage

\section{Additional Full-Scene Results}\label{supp:note5}

We show additional results on the distance map and signal recovery from simulations on rendered transients containing multi-path reflections from the iToF2dToF dataset and simulated transients using RGBD ground-truth scenes using the Middlebury and NYUv2 datasets.
We keep all the simulation parameters consistent with Suppl.~Fig.~\ref{fig:fullscene-itof2dtof} in the main text for the iToF2dToF dataset. For the Middlebury and NYUv2 datasets, we use 32-bin EDH and simulate single-bounce transients using the RGB-D data frames. For the NYU dataset, we use a $100$ns laser pulse with a FWHM = 2.5ns, and we set the deadtime at 0 and 75ns. These settings demonstrate the performance of our method under more ideal scenarios without any multi-path interference or non-Gaussian laser pulse.

\begin{figure*}[!th]
    \centering
    \includegraphics[width=0.98\linewidth]{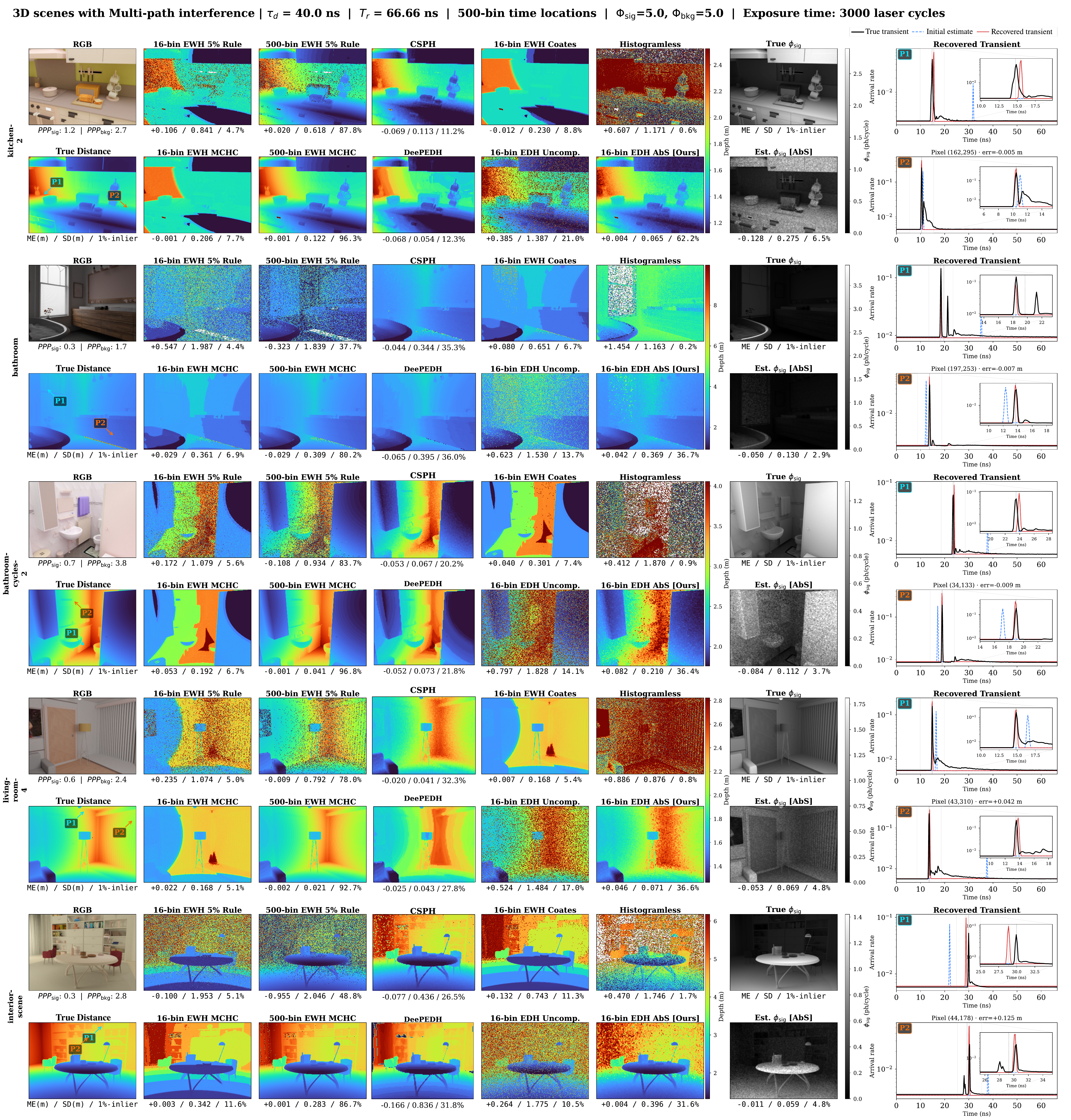}
    \caption{\textbf{Results on 3D scenes with multi-path interference from iToF2dToF dataset} We compare our AbS method with standard baselines using 16-bin Equal Width Histograms (EWH) and a histogramless approach. Quantitative results—Mean Error (ME, in meters), Standard Deviation (SD, in meters), and 1\% inlier rate (1\%in)—are reported beneath each result. As shown, our proposed method significantly reduces distance artifacts and outperforms the baselines. We also show results for signal flux estimation and distance estimated from 5\% rule method and MCHC method when operated with high-resolution EWH. The CSPH and DeePEDH show consistent negative bias and lower 1\% inliers.  Additionally, we show recovered transients for some selected pixels from each scene, demonstrating how, despite the presence of multi-path interference, our method is able to converge closer to the largest peak in most scenarios.}
    \label{fig:fullscene-itof2dtof-supp}
\end{figure*}

In the results shown below, in addition to the distance map reconstructions, we also show the signal estimates displayed as a monochrome intensity image. We also show results for 5 \% rule and the MCHC rule using 500-bin EWH (without the resource limitations). The results are shown in Suppl. Fig.~\ref{fig:fullscene-itof2dtof-supp}.

\begin{figure}[!ht]
    \centering
    \includegraphics[width=1\linewidth]{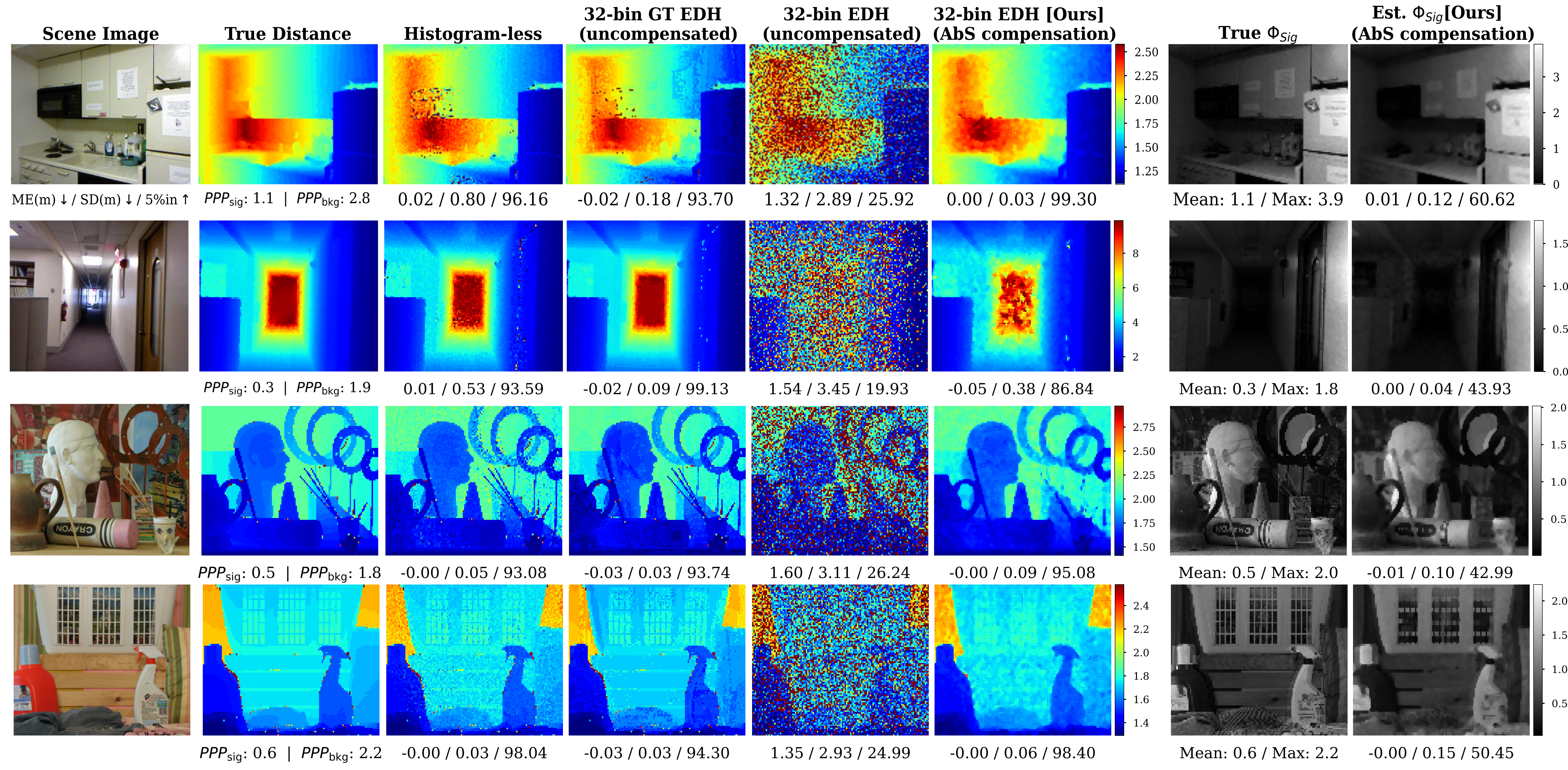}
    \caption{\textbf{Full scene results for scenes from NYUv2 and Middlebury datasets with 0 dead-time.}
    Observe that all methods provide reliable distance estimates, with the only source of noise being Poisson noise.}
    \label{fig:full_scene_no_deadtime}
\end{figure}

Suppl. Fig.~\ref{fig:full_scene_no_deadtime} shows the results with zero dead-time.
The detected photon streams do not suffer from dead-time distortions so the compared methods for distance estimation perform quite well, and the only source of noise is the Poisson noise present in the photon measurements.
We compare the distance map reconstructions of our AbS pipeline with the histogramless method of Tontini \emph{et al.}, the ground-truth (GT) 32-bin EDH quantiles of the actual perceived transients at each pixel, and the EDH measurements without any compensation.
These baseline EDH methods use the midpoint of the narrowest ED bin as an estimate of the pixel's distance.
The GT EDH can be thought of as an ``oracle'' that has access to the entire perceived transients at each pixel, and would require high in-pixel memory in practice.

\begin{figure}[!ht]
    \centering
    \includegraphics[width=1\linewidth]{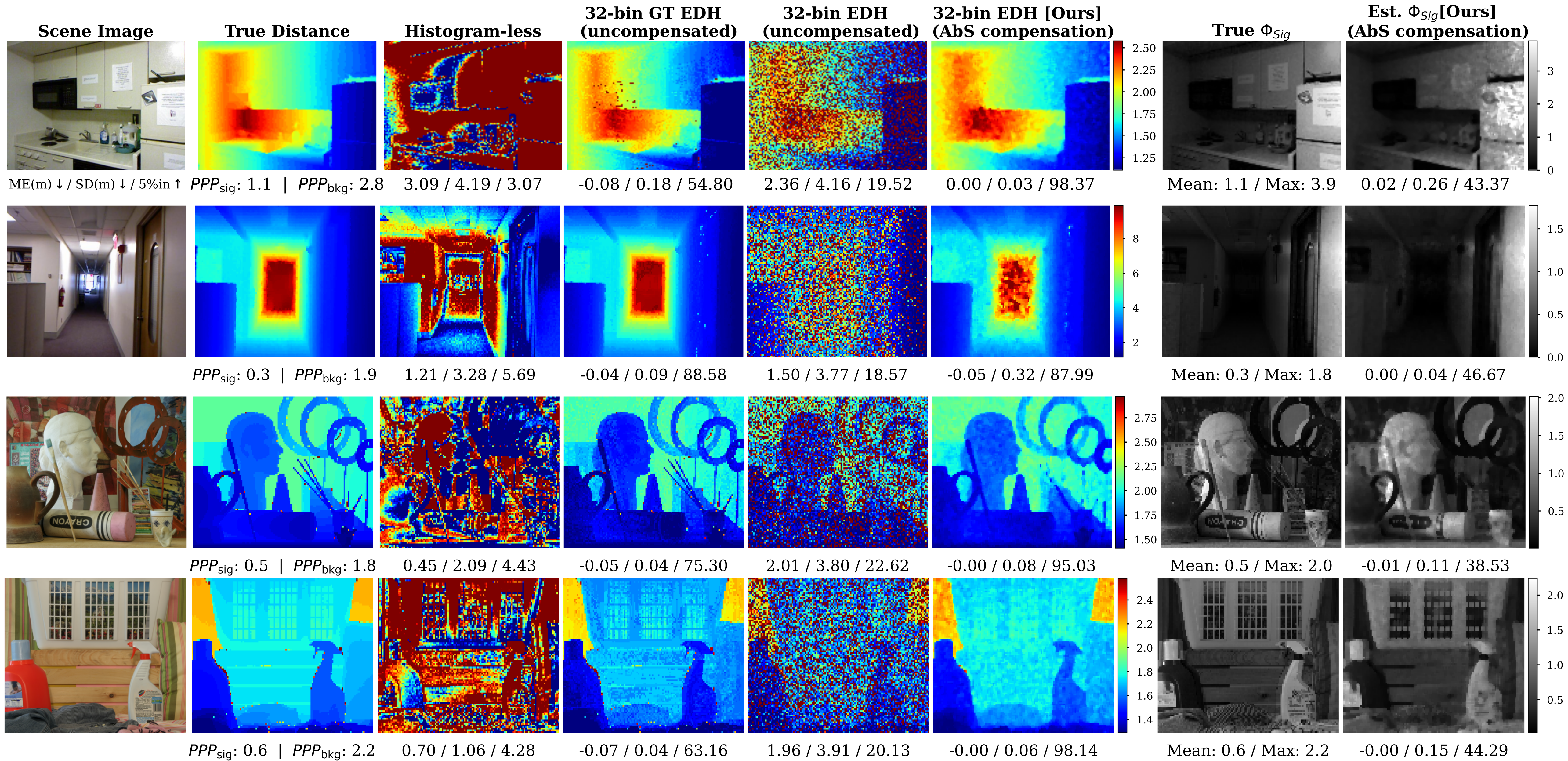}
    \caption{\textbf{Full scene results for scenes from NYUv2 and Middlebury datasets for dead-time = 75 ns.}
    The longer dead-time and high flux conditions cause strong pile-up artifacts that cannot be addressed using the histogram-less method.
    Even an ``oracle'' EDH that has access to ground truth (GT) perceived transient suffers from within-peak pile-up distortions.
    Our AbS compensation provides reliable distance and signal strength estimates even in this high-flux pile-up regime.}
    \label{fig:full_scene_75ns_deadtime}
\end{figure}

In Suppl. Fig.~\ref{fig:full_scene_75ns_deadtime}, we show simulated results when the dead-time is 75ns. 
Observe that in these high-flux conditions, the photon streams are so severely distorted that the histogram-less method fails to recover a useful distance map for any of the scenes. 
Our method continues to provide high-quality distance and signal estimates, even under such strong illumination conditions, showing the efficacy of our AbS approach.
We also note that the distance maps obtained from an ``oracle'' GT EDH look visually clean, but when compared with the results with zero dead-time, they do suffer from larger mean and inlier errors.
These larger errors are indicators of within-peak pile-up, where the peak of the laser appears shifted slightly earlier than its true position.

\clearpage

\section{Additional Calibration Robustness Results}\label{supp:note6}

\begin{figure}[!ht]
    \centering
    \includegraphics[width=1\linewidth]{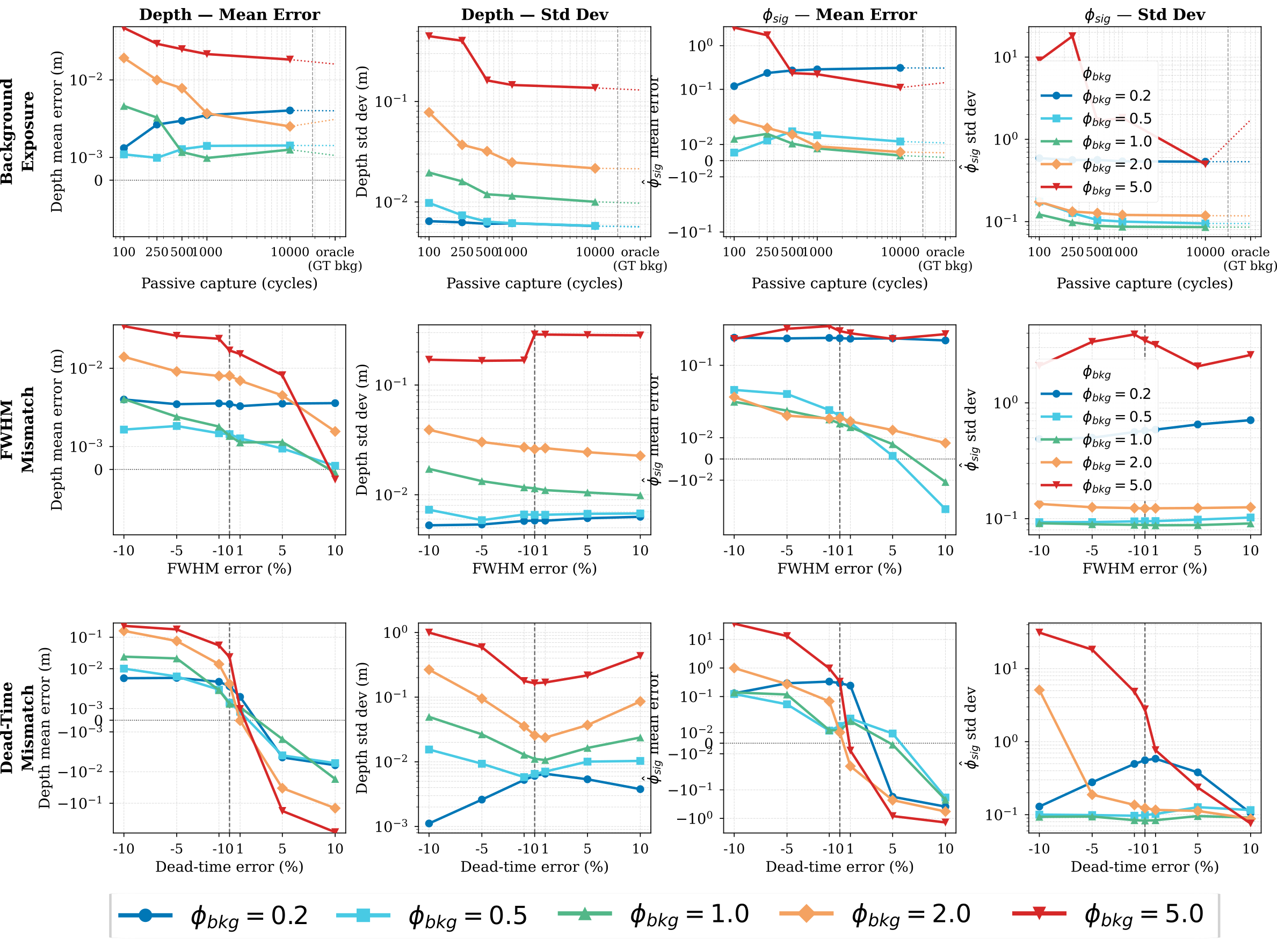}
    \caption{\textbf{Robustness of our AbS pipeline to calibration errors and model mismatch.} Our AbS pipeline estimates the scene distance and signal flux by assuming known calibrated values of \phibkg, laser pulse FWHM, and $\tau_d$. The results demonstrate a gradual increase in Mean Error (first column), standard deviation (second column) in distance estimates and in signal estimates (column 3 and 4), with increasing calibration errors in the background estimate (top row), pulse FWHM (middle row), and $\tau_d$ (bottom row). The smooth trends across five different illumination conditions demonstrate that our AbS estimator does not break under mismatch in the fixed parameters and thus works in other than ideal calibration conditions.}
    \label{fig:supp-robustness-results}
\end{figure}

\clearpage \newpage
\section{Hardware Emulation Results}\label{supp:note7}

We use a publicly available high-flux single-photon LiDAR (SPL) hardware dataset [Suppl. Ref. \ref{ref_rapp_data}] to test our AbS pipeline.
This dataset consists of long-exposure-time captures of raw photon streams using a single-pixel SPAD pixel in free-running mode.
We use the high-temporal resolution photon timestamp streams to generate the ground truth perceived transients by constructing a sum-normalized equi-width histogram of the high-resolution photon detection timestamps.
We use the MATLAB code provided with the high-flux SPL dataset to estimate the true underlying signal parameters (laser pulse FWHM and background strengths).
We use these parameters in our analytical forward model to  generate binner stationary distributions $\mathcal{D}_q$ for 31 different quantiles.
We draw random samples from these stationary distributions to emulate  32-bin EDH measurements.

We run the single-pixel AbS pipeline on these EDH measurements to estimate signal and distance.
Plugging these estimates back into our forward model we generate an estimate of the perceived transient.

\begin{figure}[!ht]
    \centering
    \includegraphics[width=0.8\linewidth]{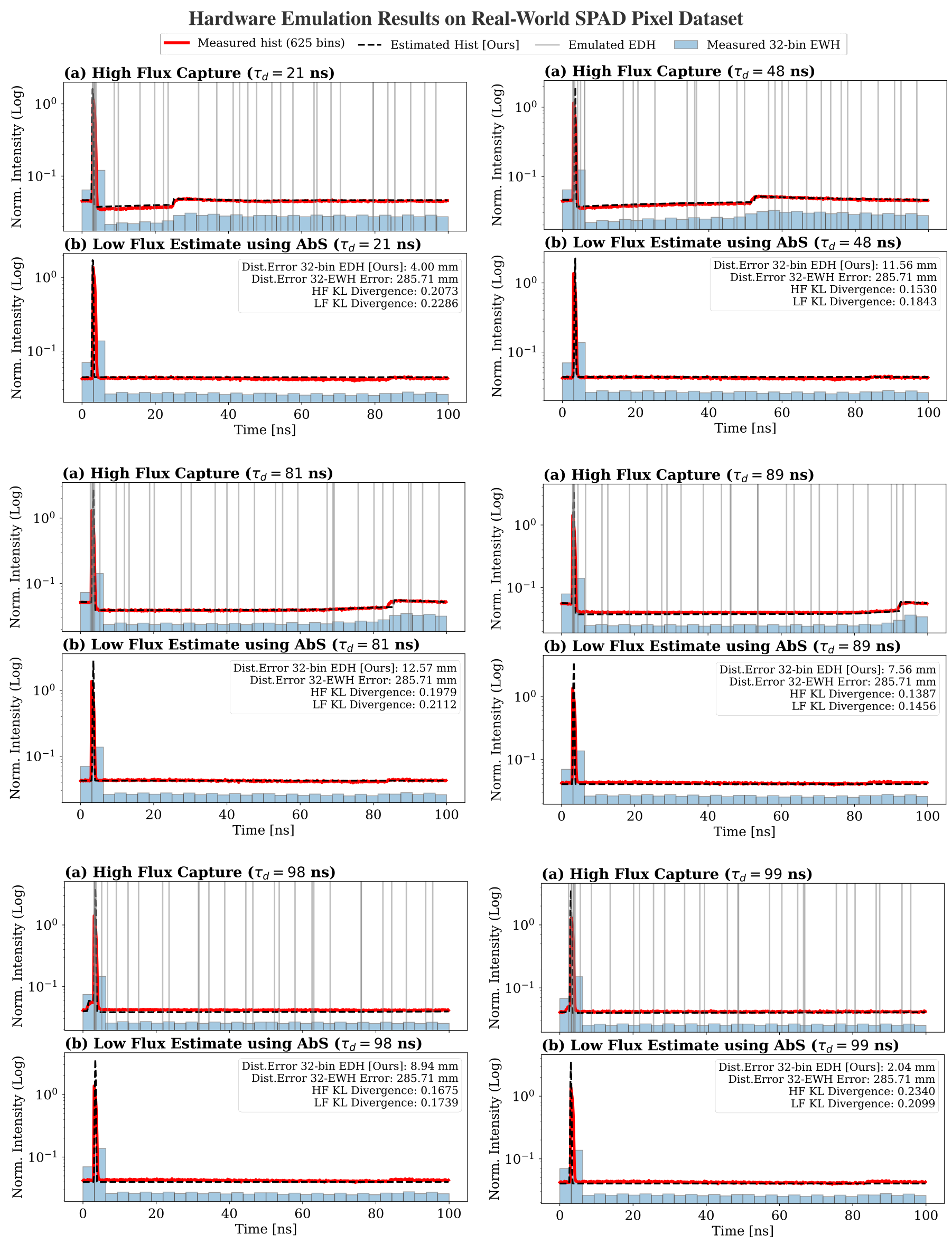}
    \caption{\textbf{Hardware emulation results for real-world SPAD pixel dataset.}
    We show results of transient reconstruction using signal and distance values estimated with our AbS pipeline on real-world photon data streams.
    Observe that there is good agreement between the actual perceived transient and the estimate over a wide range of dead-times, in both low flux (LF) and high flux (HF) conditions.
    In high-flux conditions, our method reliably predicts the within-peak pile-up and also the ``peak shadow'' i.e. the slight dip in the transient's shape that immediately follows the peak location, and spans a variable number of time locations depending on the dead-time.}
    \label{fig:hardware-results}
\end{figure}

Results shown in Suppl. Fig.~\ref{fig:hardware-results} shows good agreement between the actual perceived transient shapes and the estimated perceived transients (based on our AbS signal and distance estimates).
These results are shown for six different dead-time values (0, 21, 48,  81, 89, and 98 ns) in low-flux and high-flux conditions.
Observe the within-peak pile-up and a peak-shadow is visible in the high-flux conditions.
Our method reliably tracks the perceived transient shape in both flux conditions.

\clearpage
\newpage
\section*{Supplementary References}
\renewcommand*\labelenumi{[\theenumi]}
\begin{enumerate}
    \item Rapp et al., ``Dead-time compensation for high-flux ranging.'' \emph{IEEE Trans. Sig. Proc.}, 2019. \label{ref_rapp}
    \item G. Grimmett, D. Stirzaker, ``\emph{Probability and Random Processes.}``
      Oxford, UK: Oxford University Press, 2001, pp.~227. \label{ref_grimmett}
    \item Rapp et al., ``High-flux single-photon LiDAR.`` \emph{Optica}, 2021. 
    \label{ref_rapp_data}
\end{enumerate}

\clearpage\newpage
\newpage

\end{document}